\documentclass[lettersize,journal]{IEEEtran}
\usepackage{amsmath,amsfonts}
\usepackage{algorithm,algorithmic}
\usepackage{array}
\usepackage[caption=false,font=normalsize,labelfont=sf,textfont=sf]{subfig}
\usepackage{textcomp}
\usepackage{stfloats}
\usepackage{url}
\usepackage{verbatim}
\usepackage{graphicx}
\usepackage{cite}
\usepackage{multicol} 
\usepackage{placeins} 
\usepackage{booktabs} 
\usepackage{multirow} 
\usepackage{placeins} 
\PassOptionsToPackage{hidelinks}{hyperref} 
\usepackage{orcidlink} 

\newcommand{\argmin}{\mathop{\rm arg~min}\limits} 

\begin{document}

\title{Quaternion-Averaging-Based Adaptive Complementary Filter for Pedestrian Dead Reckoning With a Foot-Mounted AHRS}
\author{Shunsei Yamagishi\orcidlink{0000-0003-2480-3185}, Lei Jing\orcidlink{0000-0002-1181-2536} \IEEEmembership{Member, IEEE}
\thanks{This work was supported by NEDO Intensive Support for Young Promising Researchers Grant 21502121-0, JSPS KAKENHI Grant Number 26K02950, Collaborative Research with Toyota Motor Corporation, and JKA and its promotion funds from KEIRIN RACE.}
\thanks{The authors are with Graduate School of Computer Science and Engineering, The University of Aizu, Aizuwakamatsu, Fukushima 965-8580, Japan (e-mail: d8271107@u-aizu.ac.jp; leijing@u-aizu.ac.jp).}}

\maketitle

\begin{abstract}
Pedestrian Dead Reckoning (PDR) can be applied to indoor navigation systems. GPS suffers from signal degradation due to roofs and high-rise buildings, whereas PDR can estimate positions without being affected by such signal degradation. The accuracy of a foot-mounted AHRS(Attitude
and Heading Reference System)-based PDR depends on the accuracy of the attitude estimation algorithm used in the AHRS.
In this article, a Quaternion-Averaging-Based Adaptive Complementary Filter (QAACF) for PDR with a foot-mounted AHRS is proposed to improve estimation accuracy while reducing computational cost. QAACF fuses a quaternion derived from angular velocity with quaternions derived from acceleration and magnetic field measurements using Markley’s quaternion averaging, which combines two quaternions more rigorously than linear interpolation. In addition, QAACF adaptively adjusts the weights of angular velocity, acceleration, and magnetic field measurements according to gait phases and the level of magnetic disturbances.
Experimental results showed that the proposed QAACF achieves low Root Mean Square Errors (RMSEs) compared to existing attitude estimation filters while requiring lower computational cost than Kalman filters.
\end{abstract}

\begin{IEEEkeywords}
Quaternion, Quaternion averaging, Complementary filter, Kalman filter, Pedestrian Dead Reckoning, IMU sensor, AHRS, MARG Sensor.
\end{IEEEkeywords}

\section{Introduction}
\label{sec:introduction}
\IEEEPARstart{T}{he} Pedestrian Dead Reckoning (PDR) is a method for estimating a pedestrian’s walking trajectory by accumulating position changes over short periods using inertial and magnetic field data.\par
PDR has the advantage of not relying on external signals and is therefore suitable for indoor navigation systems, unlike the Global Positioning System (GPS). GPS suffers from signal degradation due to roofs and high-rise buildings, whereas PDR can estimate positions without being affected by such degradation. However, PDR suffers from accumulated estimation errors caused by acceleration and gyroscope sensor drift. In addition, AHRS-based PDR is affected by magnetic disturbances generated by electronic devices surrounding the measurement area.\par
Liu et al. \cite{a1} proposed a high-performance PDR algorithm that estimates accelerometer bias by assuming that the accelerometer bias in the body frame is constant and decomposing the attitude matrix. Kuang et al. \cite{a2} proposed a PDR method using MARG (Magnetic, Angular Rate, and Gravity) sensors attached to the foot and waist, which improves positioning accuracy by correcting the waist inertial navigation system (INS) using the foot-mounted MARG sensor and improving heading accuracy using magnetic field vector constraints. Bai et al. \cite{a3} presented a robust PDR method using a waist-mounted IMU sensor by fusing the adaptive time thresholding algorithm based on the estimated gait period (ATT-EGP), which dynamically adjusts thresholds, and an improved heuristic drift elimination (HDE) algorithm, which corrects heading estimation by recognizing motion states through turning detection. Yoshida \cite{a8} proposed an improved PDR method that classifies walking models, including straight walking, right walking, left walking, and stopping, and classifies gait phases into stance and swing using classifiers based on accelerometer and gyroscope data. Lin et al. \cite{a9} proposed a PDR method based on an array of four MEMS IMUs that reduces the horizontal positioning error by deterministic error calibration and weighted IMU data fusion using bias instability coefficients obtained from Allan variance analysis. Zhang et al. \cite{a10} proposed a PDR method for smartphones that sets optimal PDR parameters using the Arithmetic Optimization Algorithm (AOA). \par
Recently, PDR algorithms based on factor graph optimization (FGO) and deep learning have been proposed. Yue et al. \cite{a4} improved PDR accuracy by using FGO to suppress heading errors and a residual attention neural network (RANN) to estimate step length. Also, Dou et al. \cite{a5} proposed a PDR method using dual foot-mounted IMUs that incorporates stride constraints into FGO. Tu et al. \cite{a6} proposed a robust and high-accuracy dual-foot PDR method based on FGO with graduated nonconvexity (GNC), which fuses IMU measurements and interfoot distance measured from ultrasonic ranging sensors. Wang et al. \cite{a7} proposed a robust PDR method for handling magnetic disturbances by converting magnetic field data into two-dimensional images using continuous wavelet transform (CWT), applying convolutional neural networks (CNNs) for magnetic disturbance classification, and using deep deterministic policy gradient (DDPG) to adaptively fuse headings obtained from magnetometer and gyroscope data.\par
The accuracy of PDR is affected by the accuracy of the attitude estimator in an Attitude and Heading Reference System (AHRS). The attitude estimator is devided into two filters, that is, the Kalman filter and the complementary filter.\par
A Kalman filters (KFs) estimate attitude based on a statistical model under the assumption that the process and observation noises follow Gaussian distributions. Guo et al. \cite{b1} proposed a fast Kalman filter (FKF) using a computationally efficient quaternion observation model derived from accelerometer and magnetometer measurements. Dai et al. \cite{b2} proposed a Lightweight quaternion-based Extended Kalman Filter (LEKF) by simplifying the observation model using only four elements of the direction cosine matrix (DCM). Rong et al. \cite{a12} proposed a gain-regulated extended Kalman filter (GREKF), which suppresses the effect of magnetic disturbances on roll and pitch estimation by applying a diagonalizable regulation matrix (DRM) to the Kalman gain matrix. 
Shi et al. \cite{b28} proposed an attitude estimator that fuses the quaternion estimator (QUEST) \cite{b33} and the factored quaternion algorithm (FQA) \cite{b34} using linear interpolation with an adaptive interpolation factor and a Kalman filter.\par
In theory, the sigma-point Kalman filters (SPKFs) including cubature Kalman filter (CKF) can achieve higher accuracy than the extended Kalman filter (EKF), which linearizes nonlinear functions to first order \cite{b14}. The CKF that handles non-linear problems was proposed by Arasaratnam and Haykin \cite{b19}. Geng et al. \cite{b14} proposed an adaptive cubature Kalman filter (ACKF) that improves the estimation accuracy of roll, pitch, and yaw angles by updating the noise covariance matrices using two memory-weighted methods that give greater weight to recent measurements and less weight to past measurements. Yamagishi et al. \cite{a13} proposed a Kaisoku Cubature Kalman filter (KCKF) with lower computational cost by simplifying prediction equations. However, Kalman filters have high computational costs because they require matrix operations such as matrix inversion and matrix multiplication.\par
In contrast, complementary filters (CFs) have lower computational costs; however, their heading estimation accuracy is generally inferior to that of Kalman filters, especially under dynamic conditions \cite{a11}. Valenti et al. \cite{a15} proposed an adaptive-gain CF that prevents magnetic disturbances from affecting the roll and pitch estimates by separately estimating the tilt and heading quaternions. Wu et al. \cite{b6} proposed a fast complementary filter (FCF) that eliminates iterative computation by obtaining a stable solution to a linear system. Rong et al. \cite{b24} proposed an Euler-angle-based complementary filter (EuCF) that prevents magnetic disturbances from affecting roll and pitch estimation by converting quaternions into Euler angles and estimating roll, pitch, and yaw angles individually. Also, Rong et al. \cite{b27} presented a time-efficient complementary Kalman gain filter (TCF) that achieves accuracy close to that of the EKF while reducing computational cost by approximating the EKF gain matrix and avoiding costly matrix computations such as inverse matrix calculations and error covariance matrix updates. Sever et al. \cite{b30} proposed a gradient-descent-based CF that automatically determines the number of iterations using a termination criterion, thereby reducing computational cost. Kopecki et al. \cite{a14} proposed a method for optimizing the time constant of a CF using calibration data.\par
The aim of this research is to design a complementary filter (CF) that achieves higher accuracy with lower computational cost than the Kalman filter. To this end, Markley's quaternion averaging method \cite{a16} is incorporated into the proposed CF to combine two quaternions more rigorously than conventional linear interpolation. In previous studies, linear interpolation (LERP) \cite{a15,b6,b30} and spherical linear interpolation (SLERP) \cite{a15} have been used for quaternion fusion. However, LERP is not suitable for interpolating unit quaternions on the four-dimensional unit sphere $\boldsymbol{S}^{3}=\{\boldsymbol{q}\in\mathbb{R}^{4}\mid|\boldsymbol{q}|=1\}$. Moreover, SLERP alone cannot resolve the problem of the fact that the quaternions $\pm\boldsymbol{q}$ represent the same attitude. Furthermore, to improve the accuracy of attitude estimation and PDR, adaptive weights are designed based on the human gait cycle.\par
The main contributions of this paper are summarized as follows.
\begin{itemize}
    \item A low-computational-cost and high-accuracy complementary filter (CF), named the Quaternion-Averaging-Based Adaptive Complementary Filter (QAACF), is proposed.
    \item The proposed QAACF is derived from Markley's quaternion averaging method \cite{a16} by solving the characteristic equation and obtaining the maximum eigenvalue and its corresponding eigenvector, enabling two quaternions to be combined more rigorously than with conventional linear interpolation.
    \item The adaptive weights for the proposed QAACF are designed based on the human gait cycle and magnetic disturbances to improve estimation accuracy.
    \item The proposed QAACF is compared with existing KFs and CFs through controlled experiments using the root mean square errors (RMSEs) of the estimated Euler angles.
    \item The accuracy of PDR using the proposed QAACF is compared with that of PDR using existing attitude estimation algorithms.
    \item The computational cost of the proposed QAACF is compared with those of existing attitude estimation algorithms in two different computing environments: a high-performance computing environment and a low-cost computing environment.
\end{itemize}

\allowdisplaybreaks

\section{Methods}
\subsection{Overview of the Proposed QAACF}
\begin{figure*}[htbp]
    \centering
    \includegraphics[width=120mm]{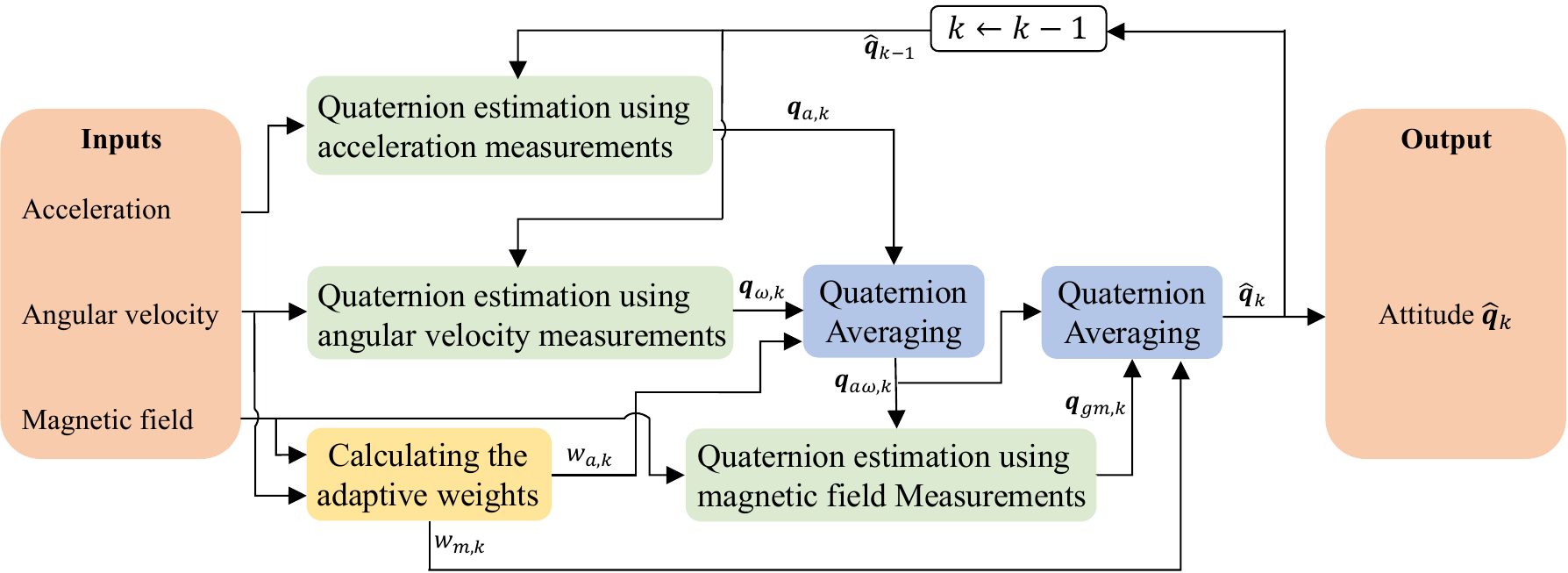}
    \caption{Overview of the Proposed QAACF}
    \label{fig:QAACF}
\end{figure*}

An overview of the proposed QAACF is shown in Fig. \ref{fig:QAACF}. The algorithmic structure of Wu's FCF \cite{b6} was used as a reference in designing the QAACF. First, the quaternion $\boldsymbol{q}_{a,k}$ is estimated using acceleration measurements, and the quaternion $\boldsymbol{q}_{\omega,k}$ is estimated using angular velocity measurements. Next, the adaptive weights $w_{a,k}$ and $w_{m,k}$ are computed. The quaternions $\boldsymbol{q}_{a,k}$ and $\boldsymbol{q}_{\omega,k}$ are then fused using Markley's quaternion averaging method to obtain the quaternion $\boldsymbol{q}_{a\omega,k}$. Subsequently, the quaternion $\boldsymbol{q}_{gm,k}$ is estimated using magnetic field measurements. Finally, $\boldsymbol{q}_{a\omega,k}$ and $\boldsymbol{q}_{gm,k}$ are fused using Markley's quaternion averaging method to obtain the attitude estimate $\hat{\boldsymbol{q}}_k$.

\subsection{Quaternion Estimation Using Angular Velocity Measurements}
The quaternion is estimated using angular velocity. The quaternion differential equation is given by (\ref{eq:dif_eq:quaternion}) \cite{a17}.
\begin{align}
    \label{eq:dif_eq:quaternion}
    \frac{d\boldsymbol{q}_k}{dt}=\frac{1}{2}\boldsymbol{q}_k\otimes
    \begin{bmatrix}
        0\\
        \boldsymbol{\omega}_k
    \end{bmatrix}
\end{align}
$\boldsymbol{q}_k=
\begin{bmatrix}
    q_{0,k} & q_{1,k} & q_{2,k} & q_{3,k}
\end{bmatrix}^T
, \boldsymbol{\omega}_k=
\begin{bmatrix}
    \omega_{x,k} & \omega_{y,k} & \omega_{z,k}
\end{bmatrix}^T$
denote the quaternion and the angular velocity at the $k$-th time step, respectively. Here, $\otimes$ denotes the Hamilton product. By applying the forward Euler method to (\ref{eq:dif_eq:quaternion}), the numerical solution is obtained as
\begin{align}
    \label{eq:state model}
    \boldsymbol{q}_k=\biggl(\boldsymbol{I}_{4\times4}+\frac{\Delta t}{2}[\boldsymbol{\Omega}_k\times]\biggr)\boldsymbol{q}_{k-1}
\end{align}
where $\boldsymbol{I}_{n\times n}$ and $\Delta t$ denote the $n\times n$ identity matrix and the sampling period, respectively. The matrix $[\boldsymbol{\Omega}_k\times]$ is defined as follows.
\begin{equation}
    \label{eq:Omega_k}
    [\boldsymbol{\Omega}_k\times]=
    \begin{bmatrix}
    0 & -\omega_{x,k} & -\omega_{y,k} & -\omega_{z,k} \\
    \omega_{x,k} & 0 & \omega_{z,k} & -\omega_{y,k} \\
    \omega_{y,k} & -\omega_{z,k} & 0 & \omega_{x,k} \\
    \omega_{z,k} & \omega_{y,k} & -\omega_{x,k} & 0
    \end{bmatrix}
\end{equation}

\subsection{Quaternion Estimation Using Acceleration Measurements}
The quaternion is estimated from the acceleration vector using the gradient-descent algorithm \cite{b4, b13, b30}. To estimate the quaternion from the acceleration vector, the following cost function, $\mathcal{L}_g(\hat{\boldsymbol{q}}_k)$, is minimized \cite{b4}.
\begin{align}
    \label{eq:cost_func}
    \min_{\hat{\boldsymbol{q}}_k\in\mathbb{R}^4}\mathcal{L}_g(\hat{\boldsymbol{q}}_k)=\min_{\hat{\boldsymbol{q}}_k\in\mathbb{R}^4}\frac{1}{2}\boldsymbol{f}_g(\hat{\boldsymbol{q}}_k)^T\boldsymbol{f}_g(\hat{\boldsymbol{q}}_k)
\end{align}
The residual vector $\boldsymbol{f}_g(\hat{\boldsymbol{q}}_k)$ is defined as follows \cite{b4}.
\begin{align}
    \boldsymbol{f}_g(\hat{\boldsymbol{q}}_k)&=\boldsymbol{C}_G^S(\hat{\boldsymbol{q}}_k)\boldsymbol{a}^r-\boldsymbol{a}_k\\
    &=
    \begin{bmatrix}
        2(q_{1,k}q_{3,k}-q_{0,k}q_{2,k})-a_{x,k}\\
        2(q_{0,k}q_{1,k}+q_{2,k}q_{3,k})-a_{y,k}\\
        2(\frac{1}{2}-q_{1,k}^2-q_{2,k}^2)-a_{z,k}
    \end{bmatrix}
\end{align} 
$\boldsymbol{a}^r=
\begin{bmatrix}
    0 & 0 & 1
\end{bmatrix}^T
, \boldsymbol{a}_k=
\begin{bmatrix}
    a_{x,k} & a_{y,k} & a_{z,k}
\end{bmatrix}^T$ represent the gravity reference vector and the measured acceleration vector, respectively. $\boldsymbol{C}_G^S(\hat{\boldsymbol{q}}_k)$ denotes the direction cosine matrix (DCM) corresponding to $\hat{\boldsymbol{q}}_k$, which transforms a vector from the global coordinate system to the sensor coordinate system, and is defined by (\ref{eq:DCM}).
\begin{figure*}
\begin{equation}
    \label{eq:DCM}
    \boldsymbol{C}^S_G(\boldsymbol{q}_k)=
    \begin{bmatrix}
        q_{0,k}^2+q_{1,k}^2-q_{2,k}^2-q_{3,k}^2 & 2(q_{1,k}q_{2,k}+q_{0,k}q_{3,k}) & 2(q_{1,k}q_{3,k}-q_{0,k}q_{2,k})\\
        2(q_{1,k}q_{2,k}-q_{0,k}q_{3,k}) & q_{0,k}^2-q_{1,k}^2+q_{2,k}^2-q_{3,k}^2 & 2(q_{2,k}q_{3,k}+q_{0,k}q_{1,k})\\
        2(q_{1,k}q_{3,k}+q_{0,k}q_{2,k}) & 2(q_{2,k}q_{3,k}-q_{0,k}q_{1,k}) & q_{0,k}^2-q_{1,k}^2-q_{2,k}^2+q_{3,k}^2
    \end{bmatrix}
\end{equation}
\end{figure*} 
The gradient of the cost function in (\ref{eq:cost_func}) is given by the following expression \cite{b4}.
\begin{align}
    \nabla\mathcal{L}_g(\hat{\boldsymbol{q}}_k)=\boldsymbol{J}_g(\hat{\boldsymbol{q}}_k)^T\boldsymbol{f}_g(\hat{\boldsymbol{q}}_k)
\end{align}
The gradient operator $\nabla$ is defined as:
\begin{align}
    \nabla=
    \begin{bmatrix}
    \frac{\partial}{\partial q_0} & \frac{\partial}{\partial q_1} & \frac{\partial}{\partial q_2} & \frac{\partial}{\partial q_3}
\end{bmatrix}^T
\end{align}
$\boldsymbol{J}_g(\hat{\boldsymbol{q}}_k)$ is the Jacobian matrix of $\boldsymbol{f}_g(\hat{\boldsymbol{q}}_k)$, which is given by \cite{b4}:
\begin{align}
    \boldsymbol{J}_g&=\left.\frac{\partial\boldsymbol{f}_g(\boldsymbol{q})}{\partial\boldsymbol{q}}\right|_{\boldsymbol{q}=\hat{\boldsymbol{q}}_k}\\
    &=
    \begin{bmatrix}
        -2q_{2,k} & 2q_{3,k} & -2q_{0,k} & 2q_{1,k}\\
        2q_{1,k} & 2q_{0,k} & 2q_{3,k} & 2q_{2,k}\\
        0 & -4q_{1,k} & -4q_{2,k} & 0
    \end{bmatrix}
\end{align}
Finally, the quaternion estimated from the acceleration vector is updated according to (\ref{eq:update_acc_quat}).
\begin{align}
    \label{eq:update_acc_quat}
    \boldsymbol{q}_{a,k}=\hat{\boldsymbol{q}}_{k-1}-\mu\nabla\mathcal{L}_g(\hat{\boldsymbol{q}}_{k-1})
\end{align}
$\mu$ is the step size of the gradient-descent algorithm.

\subsection{Quaternion Estimation Using Magnetic Field Measurements}
The quaternion is estimated using the magnetic field based on Markley's optimization method \cite{a18}. The quaternion is calculated from the magnetic field and gravity vectors according to the following equations \cite{b6,a18}.
\begin{align}
    \label{eq:calculation_of_qgm}
    \boldsymbol{q}_{gm,k}&=\frac{1}{2\sqrt{\nu(\nu+|\rho|)c}}\times
    &\begin{cases}
        \begin{bmatrix}
            (\nu+\rho)c \\
            (\nu+\rho)\boldsymbol{l}+\sigma\boldsymbol{d}
        \end{bmatrix} & \text{if $\rho\geq0$}\\
        \begin{bmatrix}
            \sigma c\\
            \sigma\boldsymbol{l}+(\nu-\rho)\boldsymbol{d}
        \end{bmatrix} & \text{if $\rho<0$}
    \end{cases}
\end{align}
The variables $\rho$, $\sigma$, and $\nu$ are defined as follows \cite{b6,a18}.
\begin{align}
    \rho&=c\{\tilde{\omega}\boldsymbol{b}_1\cdot\boldsymbol{r}_1+(1-\tilde{\omega})\boldsymbol{b}_2\cdot\boldsymbol{r}_2\}+\boldsymbol{l}\cdot\boldsymbol{u}\\
    \sigma&=\boldsymbol{d}\cdot\boldsymbol{u}\\
    \nu&=\sqrt{\rho^2+\sigma^2}
\end{align}
The variables $c$, $\boldsymbol{d}$, $\boldsymbol{l}$, and $\boldsymbol{u}$ are defined as follows \cite{b6,a18}.
\begin{align}
    c&=1+\boldsymbol{b}_\times\cdot\boldsymbol{r}_\times\\
    \boldsymbol{d}&=\boldsymbol{b}_\times+\boldsymbol{r}_\times\\
    \boldsymbol{l}&=\boldsymbol{b}_\times\times\boldsymbol{r}_\times\\
    \boldsymbol{u}&=\tilde{\omega}\boldsymbol{b}_1\times\boldsymbol{r}_1+(1-\tilde{\omega})\boldsymbol{b}_2\times\boldsymbol{r}_2
\end{align}
The unit vectors $\boldsymbol{b}_\times$ and $\boldsymbol{r}_\times$ are defined as follows \cite{b6,a18}.
\begin{align}
    \boldsymbol{r}_\times&=\frac{\boldsymbol{r}_1\times\boldsymbol{r}_2}{\|\boldsymbol{r}_1\times\boldsymbol{r}_2\|}\\
    \boldsymbol{b}_\times&=\frac{\boldsymbol{b}_1\times\boldsymbol{b}_2}{\|\boldsymbol{b}_1\times\boldsymbol{b}_2\|}
\end{align}
Finally, the vectors $\boldsymbol{b}_1$, $\boldsymbol{r}_1$, $\boldsymbol{b}_2$, and $\boldsymbol{r}_2$ are defined as follows \cite{b6}.
\begin{align}
    \boldsymbol{b}_1&=\boldsymbol{g}^S_k=\boldsymbol{C}_G^S(\boldsymbol{q}_{a\omega,k})\boldsymbol{a}^r\\
    \boldsymbol{r}_1&=\boldsymbol{a}^r\\
    \boldsymbol{b}_2&=\boldsymbol{m}_k\\
    \label{eq:mag_ref_vec}
    \boldsymbol{r}_2&=
    \begin{bmatrix}
        m_{N,k} & 0 & m_{D,k}
    \end{bmatrix}^T
\end{align}
$\boldsymbol{m}_k=
\begin{bmatrix}
    m_{x,k} & m_{y,k} & m_{z,k}
\end{bmatrix}^T$
denotes the measured magnetic field vector at the $k$-th time step. $\boldsymbol{q}_{a\omega,k}$ denotes the quaternion estimated from angular velocity and acceleration measurements. The scalars $m_{D,k}$ and $m_{N,k}$ are computed as follows \cite{b6}.
\begin{align}
    \label{eq:mD}
    m_{D,k}&=\boldsymbol{g}^S_k\cdot\boldsymbol{m}_k\\
    \label{eq:mN}
    m_{N,k}&=\sqrt{1-m_{D,k}^2}
\end{align}
The parameter $\tilde{\omega}\in[0,1]$ is the weighting coefficient in Wahba's problem \cite{b6,a18,b32}. In this paper, $\tilde{\omega}$ is set to 0.5.

\subsection{Quaternion Averaging-Based Fusion}
In this paper, a closed-form fusion formula for two quaternions is derived from Markley's quaternion averaging method \cite{a16}.
Markley's quaternion averaging method \cite{a16} is incorporated into the proposed QAACF to fuse two quaternions. Markley's quaternion averaging method \cite{a16} combines two quaternions more rigorously than conventional LERP, and handles a problem that quaternions $\pm\boldsymbol{q}$ represent the same attitude. Markley's quaternion averaging method finds the quaternion that minimizes the loss function, defined as follows (\ref{eq:def_quaternion_avg}) \cite{a16}: 
\begin{align}
    \label{eq:def_quaternion_avg}
    \overline{\boldsymbol{q}}=\argmin_{\boldsymbol{q}\in\boldsymbol{S}^3} \sum_{i=1}^nw_i\|\boldsymbol{C}_G^S(\boldsymbol{q})-\boldsymbol{C}_G^S(\boldsymbol{q}_i)\|_F^2
\end{align}
$\|\cdot\|_F$ represents the Frobenius norm. $\overline{\boldsymbol{q}}$ is obtained by finding the eigenvector corresponding to the largest eigenvalue of the following matrix $\boldsymbol{M}$ \cite{a16}.
\begin{align}
    \boldsymbol{M}=\sum_{i=1}^nw_i\boldsymbol{q}_i\boldsymbol{q}_i^T
\end{align}
In the proposed QAACF, the case of $n=2$ is considered. Accordingly, the matrix $\boldsymbol{M}$ is given by
\begin{align}
    \boldsymbol{M}=(1-w)\boldsymbol{q}_1\boldsymbol{q}_1^T+w\boldsymbol{q}_2\boldsymbol{q}_2^T
\end{align}

The eigenvector corresponding to the largest eigenvalue is obtained by solving the following characteristic equation (\ref{eq:characteristic_eq_1}).
\begin{align}
    \label{eq:characteristic_eq_1}
    \Phi_{\boldsymbol{M}}(\lambda)=0
\end{align}
Here, $\Phi_{\boldsymbol{M}}(\lambda)$ denotes the characteristic polynomial given in (\ref{eq:characteristic_polynomial}).
\begin{align}
    \label{eq:characteristic_polynomial}
    \Phi_{\boldsymbol{M}}(\lambda)=\det(\lambda\boldsymbol{I}_{4\times4}-\boldsymbol{M})
\end{align}
The derivation is given below.
\begin{align}
    \Phi_{\boldsymbol{M}}(\lambda)&=\det(\lambda\boldsymbol{I}_{4\times4}-\boldsymbol{M})\\
    &=\det\{\lambda(\boldsymbol{I}_{4\times4}-\frac{1}{\lambda}\boldsymbol{M})\}\\
    &\text{Using $\det(c\boldsymbol{A})=c^n\det(\boldsymbol{A})$, where $c\in\mathbb{R}$ and} \notag\\
    &\text{$\boldsymbol{A}\in\mathbb{R}^{n\times n}$,}\notag\\
    &=\lambda^4\det(\boldsymbol{I}_{4\times4}-\frac{1}{\lambda}\boldsymbol{M})\\
    &\text{Substituting $\boldsymbol{M}=(1-w)\boldsymbol{q}_1\boldsymbol{q}_1^T+w\boldsymbol{q}_2\boldsymbol{q}_2^T$,}\notag\\
    &=\lambda^4\det[\boldsymbol{I}_{4\times4}-\frac{1}{\lambda}\{(1-w)\boldsymbol{q}_1\boldsymbol{q}_1^T+w\boldsymbol{q}_2\boldsymbol{q}_2^T\}]\\
    &=\lambda^4\det(\boldsymbol{I}_{4\times4}\notag\\
    &\quad-
    \begin{bmatrix}
        \sqrt{\frac{1-w}{\lambda}}\boldsymbol{q}_1 & \sqrt{\frac{w}{\lambda}}\boldsymbol{q}_2
    \end{bmatrix}
    \begin{bmatrix}
        \sqrt{\frac{1-w}{\lambda}}\boldsymbol{q}_1^T\\
        \sqrt{\frac{w}{\lambda}}\boldsymbol{q}_2^T
    \end{bmatrix}
    )
\end{align}
Here, we define $\boldsymbol{U}=
\begin{bmatrix}
    \sqrt{\frac{1-w}{\lambda}}\boldsymbol{q}_1 & \sqrt{\frac{w}{\lambda}}\boldsymbol{q}_2
\end{bmatrix},
\boldsymbol{V}=
\begin{bmatrix}
    \sqrt{\frac{1-w}{\lambda}}\boldsymbol{q}_1&
    \sqrt{\frac{w}{\lambda}}\boldsymbol{q}_2
\end{bmatrix}
$. \\
Using $\det(\boldsymbol{I}_{4\times4}-\boldsymbol{U}\boldsymbol{V}^T)=\det(\boldsymbol{I}_{2\times2}-\boldsymbol{V}^T\boldsymbol{U})$ (Weinstein–Aronszajn identity), we obtain
\begin{align}
    \Phi_{\boldsymbol{M}}(\lambda)&=\lambda^4\det\Biggl(\boldsymbol{I}_{2\times2}\notag\\
    &\quad-
    \begin{bmatrix}
    \sqrt{\frac{1-w}{\lambda}}\boldsymbol{q}_1^T\\
    \sqrt{\frac{w}{\lambda}}\boldsymbol{q}_2^T
\end{bmatrix}
\begin{bmatrix}
    \sqrt{\frac{1-w}{\lambda}}\boldsymbol{q}_1 & \sqrt{\frac{w}{\lambda}}\boldsymbol{q}_2
\end{bmatrix}\Biggr)\\
&=\lambda^4\det\Biggl(\boldsymbol{I}_{2\times2}\notag\\
&\quad-
\begin{bmatrix}
    \frac{1-w}{\lambda}\boldsymbol{q}_1^T\boldsymbol{q}_1 & \frac{\sqrt{w(1-w)}}{\lambda}\boldsymbol{q}_1^T\boldsymbol{q}_2\\
    \frac{\sqrt{w(1-w)}}{\lambda}\boldsymbol{q}_2^T\boldsymbol{q}_1 &\frac{w}{\lambda}\boldsymbol{q}_2^T\boldsymbol{q}_2
\end{bmatrix}
\Biggr)\\
&=\lambda^4\det
\begin{bmatrix}
    1-\frac{1-w}{\lambda}\boldsymbol{q}_1^T\boldsymbol{q}_1 & \frac{\sqrt{w(1-w)}}{\lambda}\boldsymbol{q}_1^T\boldsymbol{q}_2\\
    \frac{\sqrt{w(1-w)}}{\lambda}\boldsymbol{q}_2^T\boldsymbol{q}_1 & 1-\frac{w}{\lambda}\boldsymbol{q}_2^T\boldsymbol{q}_2
\end{bmatrix}\\
&\text{Since $\boldsymbol{q}_i^T\boldsymbol{q}_i=\|\boldsymbol{q}_i\|^2$, $\boldsymbol{q}_i^T\boldsymbol{q}_j=\boldsymbol{q}_i\cdot\boldsymbol{q}_j$,}\notag\\
&=\lambda^4\det
\begin{bmatrix}
    1-\frac{1-w}{\lambda}\|\boldsymbol{q}_1\|^2 & \frac{\sqrt{w(1-w)}}{\lambda}\boldsymbol{q}_1\cdot\boldsymbol{q}_2\\
    \frac{\sqrt{w(1-w)}}{\lambda}\boldsymbol{q}_2\cdot\boldsymbol{q}_1 & 1-\frac{w}{\lambda}\|\boldsymbol{q}_2\|^2
\end{bmatrix}\\
&\text{Applying the commutative property of the inner }\notag\\
&\text{product,}\notag\\
&=\lambda^4\det
\begin{bmatrix}
    1-\frac{1-w}{\lambda}\|\boldsymbol{q}_1\|^2 & \frac{\sqrt{w(1-w)}}{\lambda}\boldsymbol{q}_1\cdot\boldsymbol{q}_2\\
    \frac{\sqrt{w(1-w)}}{\lambda}\boldsymbol{q}_1\cdot\boldsymbol{q}_2 & 1-\frac{w}{\lambda}\|\boldsymbol{q}_2\|^2
\end{bmatrix}\\
&\text{Using 
$
\det\begin{bmatrix}
    a & b\\
    c & d
\end{bmatrix}
=ad-bc
$,}\notag\\
&=\lambda^4\biggl[\biggl(1-\frac{1-w}{\lambda}\|\boldsymbol{q}_1\|^2\biggr)\biggl(1-\frac{w}{\lambda}\|\boldsymbol{q}_2\|^2\biggr) \notag\\
&\quad-\frac{w(1-w)}{\lambda^2}(\boldsymbol{q}_1\cdot\boldsymbol{q}_2)^2\biggr]\\
&\text{Since $\boldsymbol{q}_i$ is a unit quaternion satisfying $\|\boldsymbol{q}_i\|^2=1$,}\notag\\
&=\lambda^4\biggl[\biggl(1-\frac{1-w}{\lambda}\times1\biggr)\biggl(1-\frac{w}{\lambda}\times1\biggr) \notag\\
&\quad-\frac{w(1-w)}{\lambda^2}(\boldsymbol{q}_1\cdot\boldsymbol{q}_2)^2\biggr]\\
&\text{Using the distributive law,}\notag\\
&=\lambda^4\biggl\{1-\frac{1-w}{\lambda}-\frac{w}{\lambda}+\frac{w(1-w)}{\lambda^2}\notag\\
&\quad-\frac{w(1-w)}{\lambda^2}(\boldsymbol{q}_1\cdot\boldsymbol{q}_2)^2\biggr\}\\
&\text{Distributing $\lambda^2$, we obtain} \notag\\
&=\lambda^2\{\lambda^2-(1-w)\lambda-w\lambda+w(1-w) \notag\\
&\quad-w(1-w)(\boldsymbol{q}_1\cdot\boldsymbol{q}_2)^2\}\\
&=\lambda^2[\lambda^2-\lambda+w(1-w)\{1-(\boldsymbol{q}_1\cdot\boldsymbol{q}_2)^2\}]
\end{align}
Thus, the following characteristic equation (\ref{eq:eq:characteristic_eq_2}) is obtained.
\begin{align}
    \label{eq:eq:characteristic_eq_2}
    \lambda^2[\lambda^2-\lambda+w(1-w)\{1-(\boldsymbol{q}_1\cdot\boldsymbol{q}_2)^2\}]=0
\end{align}
By solving (\ref{eq:eq:characteristic_eq_2}), we obtain the following eigenvalues.
\begin{align}
    \lambda=0,\frac{1\pm\sqrt{1-4w(1-w)\{1-(\boldsymbol{q}_1\cdot\boldsymbol{q}_2)^2\}}}{2}
\end{align}
Therefore, the largest eigenvalue $\Lambda$ is given by (\ref{eq:largest_eigenvalue}).
\begin{align}
    \label{eq:largest_eigenvalue}
    \Lambda=\frac{1+\sqrt{1-4w(1-w)\{1-(\boldsymbol{q}_1\cdot\boldsymbol{q}_2)^2\}}}{2}
\end{align}
Next, we derive an eigenvector $\overline{\boldsymbol{q}}$ corresponding to the largest eigenvalue $\Lambda$.\\
First, we show that $\forall\boldsymbol{x}\in\mathbb{R}^4,\boldsymbol{M}\boldsymbol{x}\in\operatorname{span}\{\boldsymbol{q}_1,\boldsymbol{q}_2\}$.
\begin{align}
    \boldsymbol{M}\boldsymbol{x}&=\{(1-w)\boldsymbol{q}_1\boldsymbol{q}_1^T+w\boldsymbol{q}_2\boldsymbol{q}_2^T\}\boldsymbol{x}\\
    &=(1-w)\boldsymbol{q}_1\boldsymbol{q}_1^T\boldsymbol{x}+w\boldsymbol{q}_2\boldsymbol{q}_2^T\boldsymbol{x}\\
    &\text{Here, let $\alpha=\boldsymbol{q}_1^T\boldsymbol{x},\beta=\boldsymbol{q}_2^T\boldsymbol{x}$.}\notag\\
    \boldsymbol{M}\boldsymbol{x}&=(1-w)\alpha\boldsymbol{q}_1+w\beta\boldsymbol{q}_2
\end{align}
Thus, $\forall\boldsymbol{x},\boldsymbol{M}\boldsymbol{x}\in\operatorname{span}\{\boldsymbol{q}_1,\boldsymbol{q}_2\}$.\\
Since $\Lambda$ is the largest eigenvalue of $\boldsymbol{M}$ and $\overline{\boldsymbol{q}}$ is the corresponding eigenvector, the following equation is satisfied.
\begin{align}
    \label{eq:eigenvalue_and_eigenvector}
    \boldsymbol{M}\overline{\boldsymbol{q}}=\Lambda\overline{\boldsymbol{q}}
\end{align}
From (\ref{eq:eigenvalue_and_eigenvector}), $\Lambda\overline{\boldsymbol{q}}\in\operatorname{span}\{\boldsymbol{q}_1,\boldsymbol{q}_2\}$, because $\boldsymbol{M}\overline{\boldsymbol{q}}\in\operatorname{span}\{\boldsymbol{q}_1,\boldsymbol{q}_2\}$.
Thus, $\overline{\boldsymbol{q}}\in\operatorname{span}\{\boldsymbol{q}_1,\boldsymbol{q}_2\}$.
Therefore, we can write $\overline{\boldsymbol{q}}=a\boldsymbol{q}_1+b\boldsymbol{q}_2$.\\
Then, we consider the left-hand side of (\ref{eq:eigenvalue_and_eigenvector}).
\begin{align}
    \boldsymbol{M}\overline{\boldsymbol{q}}&=\{(1-w)\boldsymbol{q}_1\boldsymbol{q}_1^T+w\boldsymbol{q}_2\boldsymbol{q}_2^T\}(a\boldsymbol{q}_1+b\boldsymbol{q}_2)\\
    &=a(1-w)\boldsymbol{q}_1\boldsymbol{q}_1^T\boldsymbol{q}_1+wa\boldsymbol{q}_2\boldsymbol{q}_2^T\boldsymbol{q}_1\notag\\
    &\quad+(1-w)b\boldsymbol{q}_1\boldsymbol{q}_1^T\boldsymbol{q}_2+wb\boldsymbol{q}_2\boldsymbol{q}_2^T\boldsymbol{q}_2\\
    &=a(1-w)\|\boldsymbol{q}_1\|^2\boldsymbol{q}_1+wa(\boldsymbol{q}_1\cdot\boldsymbol{q}_2)\boldsymbol{q}_2 \notag\\
    &\quad+(1-w)b(\boldsymbol{q}_1\cdot\boldsymbol{q}_2)\boldsymbol{q}_1+wb\|\boldsymbol{q}_2\|^2\boldsymbol{q}_2\\
    &=a(1-w)\boldsymbol{q}_1+wa(\boldsymbol{q}_1\cdot\boldsymbol{q}_2)\boldsymbol{q}_2 \notag\\
    &\quad+(1-w)b(\boldsymbol{q}_1\cdot\boldsymbol{q}_2)\boldsymbol{q}_1+wb\boldsymbol{q}_2\\
    \label{eq:Mq=xq_1+yq_2}
    &=\{a(1-w)+(1-w)b(\boldsymbol{q}_1\cdot\boldsymbol{q}_2)\}\boldsymbol{q}_1 \notag\\
    &\quad+w\{a(\boldsymbol{q}_1\cdot\boldsymbol{q}_2)+b\}\boldsymbol{q}_2
\end{align}
Then, we consider the right-hand side of (\ref{eq:eigenvalue_and_eigenvector}).
\begin{align}
    \Lambda\overline{\boldsymbol{q}}&=\Lambda(a\boldsymbol{q}_1+b\boldsymbol{q}_2)\\
    \label{eq:lambda q=xq_1+yq_2}
    &=\Lambda a\boldsymbol{q}_1+\Lambda b\boldsymbol{q}_2
\end{align}
Comparing the coefficients in (\ref{eq:Mq=xq_1+yq_2}) and (\ref{eq:lambda q=xq_1+yq_2}), we obtain the following two equations.
\begin{align}
    \label{eq:renritu1}
    a(1-w)+(1-w)b(\boldsymbol{q}_1\cdot\boldsymbol{q}_2)=\Lambda a\\
    \label{eq:renritu2}
    w\{a(\boldsymbol{q}_1\cdot\boldsymbol{q}_2)+b\}=\Lambda b
\end{align}
From (\ref{eq:renritu2}),
\begin{align}
    wa\boldsymbol{q}_1\cdot\boldsymbol{q}_2+(w-\Lambda)b=0\\
    \Leftrightarrow w\boldsymbol{q}_1\cdot\boldsymbol{q}_2+(w-\Lambda)\frac{b}{a}=0\\
    \Leftrightarrow(\Lambda-w)\frac{b}{a}=w\boldsymbol{q}_1\cdot\boldsymbol{q}_2\\
    \Leftrightarrow\frac{b}{a}=\frac{w\boldsymbol{q}_1\cdot\boldsymbol{q}_2}{\Lambda-w}
\end{align}
Therefore,
\begin{align}
    \overline{\boldsymbol{q}}&=a\boldsymbol{q}_1+b\boldsymbol{q}_2\\
    &=a(\boldsymbol{q}_1+\frac{b}{a}\boldsymbol{q}_2)\\
    &\propto\boldsymbol{q}_1+\frac{b}{a}\boldsymbol{q}_2\\
    &=\boldsymbol{q}_1+\frac{w\boldsymbol{q}_1\cdot\boldsymbol{q}_2}{\Lambda-w}\boldsymbol{q}_2
\end{align}
Therefore, we obtain the following closed-form fusion formula for two quaternions.
\begin{align}
    \overline{\boldsymbol{q}}\propto\boldsymbol{q}_1+\frac{w\boldsymbol{q}_1\cdot\boldsymbol{q}_2}{\Lambda-w}\boldsymbol{q}_2
\end{align}

\subsection{Adaptive Weight Calculation}
\begin{figure}[htbp]
    \centering
    \includegraphics[width=90mm]{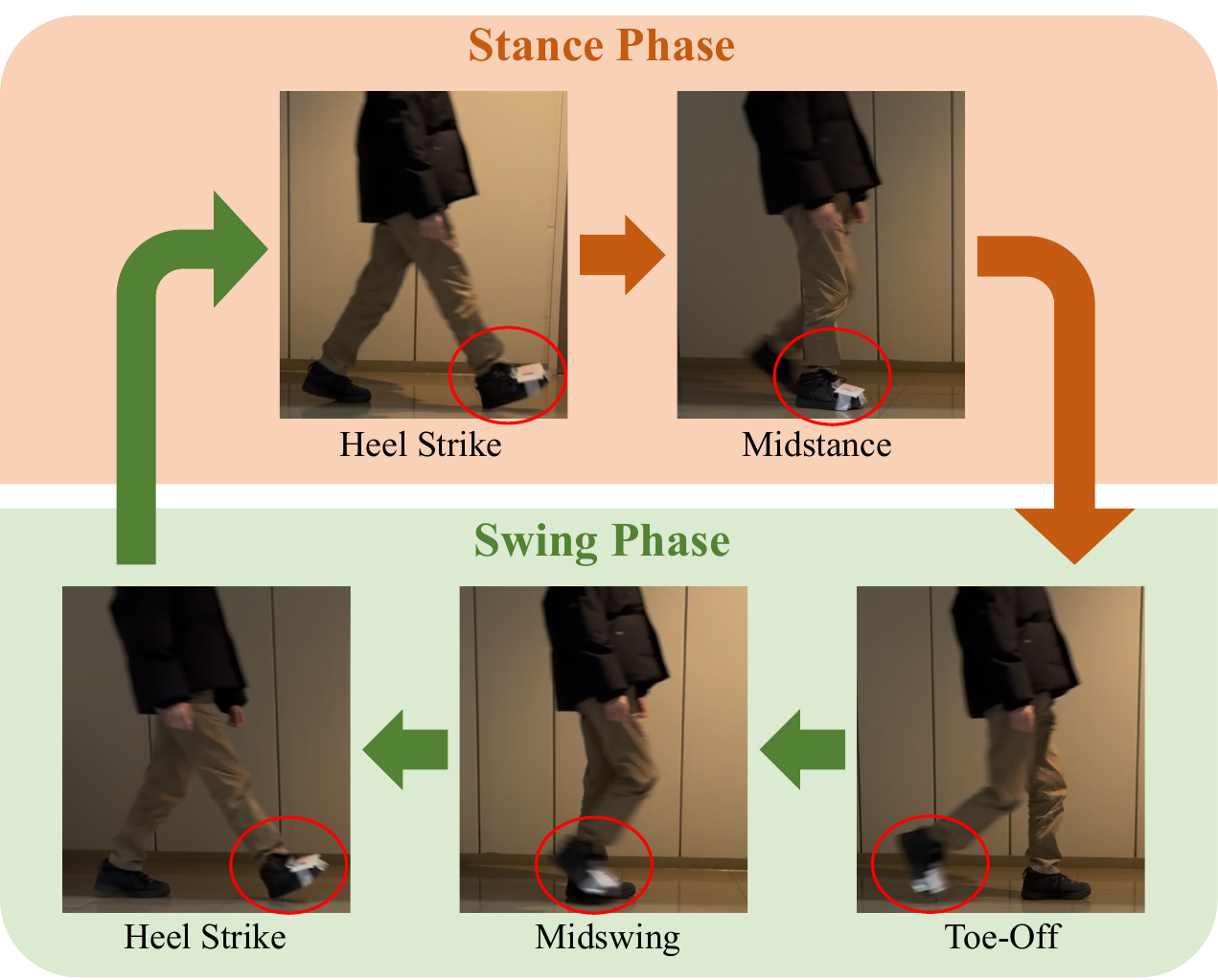}
    \caption{Definition of the gait phases according to \cite{a19}}
    \label{fig:gait_phases}
\end{figure}

The adaptive weight used in quaternion averaging is designed based on the human gait phases and magnetic disturbances. According to Sherratt \cite{a19}, the gait cycle is divided into the stance phase, which consists of heel strike and midstance, and the swing phase, which consists of toe-off, midswing, and heel strike, as shown in Fig. \ref{fig:gait_phases}.
Let the weight assigned to $\boldsymbol{q}_{\omega,k}$, estimated from the angular velocity, be $1-w_{a,k}$ and the weight assigned to $\boldsymbol{q}_{a,k}$, estimated from the acceleration, be $w_{a,k}$. Then, the matrix $\boldsymbol{M}$ used for quaternion averaging is given by
\begin{align}
    \boldsymbol{M}=(1-w_{a,k})\boldsymbol{q}_{\omega,k}\boldsymbol{q}_{\omega,k}^T+w_{a,k}\boldsymbol{q}_{a,k}\boldsymbol{q}_{a,k}^T
\end{align}
Since the foot is almost stationary and the dynamic acceleration noise is negligible during the midstance phase, $\boldsymbol{q}_{a,k}$ is more reliable, and the weight assigned to $\boldsymbol{q}_{a,k}$ should be increased. The adaptive weight $w_{a,k}$ is calculated as
\begin{align}
    w_{a,k}=\alpha_ae^{-\beta_a\|\boldsymbol{\omega}_k\|^2}
\end{align}
where $\alpha_a\in[0,1]$ and $\beta_a\in[0,\infty)$ are the parameters for determining the adaptive weight $w_{a,k}$. $\|\boldsymbol{\omega}_k\|$ denotes the magnitude of the angular velocity vector used for midstance phase detection \cite{a20}.\\
Next, let the weight assigned to $\boldsymbol{q}_{a\omega,k}$, estimated from the acceleration and angular velocity, be $1-w_{m,k}$ and the weight assigned to $\boldsymbol{q}_{gm,k}$, estimated from the magnetic field, be $w_{m,k}$. Then, the matrix $\boldsymbol{M}$ used for quaternion averaging is given by
\begin{align}
    \boldsymbol{M}=(1-w_{m,k})\boldsymbol{q}_{a\omega,k}\boldsymbol{q}_{a\omega,k}^T+w_{m,k}\boldsymbol{q}_{gm,k}\boldsymbol{q}_{gm,k}^T
\end{align}
According to (\ref{eq:mag_ref_vec}), (\ref{eq:mD}), and (\ref{eq:mN}), during the non-midstance phase, the magnetic reference vector $\boldsymbol{r}_2$ contains dynamic acceleration noise. Therefore, when the foot is not in the midstance phase or the magnetic field is disturbed, $\boldsymbol{q}_{gm,k}$ is less reliable, and its weight should be reduced. 
Hence, the adaptive weight $w_{m,k}$ is calculated as
\begin{align}
    e_{m,k}&=\frac{m_k^{abs}-m_s}{m_s}\\
    w_{m,k}&=\alpha_me^{-\beta_a\|\boldsymbol{\omega}_k\|^2}e^{-\beta_me_{m,k}^2}\\
    &=\alpha_m\exp\!\left(
-\beta_a\|\boldsymbol{\omega}_k\|^2
-\beta_me_{m,k}^2
\right)
\end{align}
where $\alpha_m\in[0,1]$ and $\beta_m\in[0,\infty)$ are the parameters for setting the adaptive weight $w_{m,k}$. Here, $m_k^{abs}$ and $m_s$ denote the magnitude of the magnetic field vector measured by the magnetometer and the magnitude of the magnetic field vector when the magnetic field is the most stable conditions, respectively.

\subsection{Summary of the Proposed QAACF Algorithm}
The pseudo-code (Algorithm \ref{alg:QAACF}: Quaternion-Averaging-Based Complementary Filter (QAACF)) presents the overall procedure of the proposed QAACF. The inputs to the QAACF are the acceleration, angular velocity, and magnetic field measured by the AHRS. The parameters required to configure the QAACF are $\mu$, $\alpha_a$, $\beta_a$, $\alpha_m$, $\beta_m$, and $m_s$. The output of the QAACF is the estimated attitude of the AHRS. Here, $t_k$ denotes the time at the $k$-th time step. For $t_k \leq 3$ seconds, the adaptive weights $w_{a,k}$ and $w_{m,k}$ are both set to 0.5 to accelerate convergence. During the initial convergence period ($t_k \leq 3$ seconds), the AHRS should remain stationary.
 
 \begin{figure}[!t]
\begin{algorithm}[H]
	\caption{Quaternion-Averaging-Based Complementary Filter (QAACF)}
	\label{alg:QAACF}
	\begin{algorithmic}[1]
	\REQUIRE $\boldsymbol{a}_k,\boldsymbol{\omega}_k,\boldsymbol{m}_k,\hat{\boldsymbol{q}}_0,\mu,\alpha_a,\beta_a,\alpha_m,\beta_m,m_s$
    \ENSURE $\hat{\boldsymbol{q}}_k$
    \FOR{k}
        \STATE $m_k^{abs} \leftarrow \|\boldsymbol{m}_k\|$
        \STATE $\boldsymbol{a}_k \leftarrow \frac{\boldsymbol{a}_k}{\|\boldsymbol{a}_k\|}$, $\boldsymbol{m}_k \leftarrow \frac{\boldsymbol{m}_k}{\|\boldsymbol{m}_k\|}$
        \STATE $\nabla\boldsymbol{f}_{g}=\boldsymbol{J}_g(\hat{\boldsymbol{q}}_{k-1})^T\boldsymbol{f}_g(\hat{\boldsymbol{q}}_{k-1},\boldsymbol{a}_k)$
        \STATE $\boldsymbol{q}_{a,k}=\hat{\boldsymbol{q}}_{k-1}-\mu\nabla\boldsymbol{f}_g$
        \STATE $\boldsymbol{q}_{a,k} \leftarrow \frac{\boldsymbol{q}_{a,k}}{\|\boldsymbol{q}_{a,k}\|}$
        \STATE $\boldsymbol{q}_{\omega,k}=\bigl(\boldsymbol{I}_{4\times4}+\frac{\Delta t}{2}[\boldsymbol{\Omega}_k\times]\bigr)\hat{\boldsymbol{q}}_{k-1}$
        \STATE $\boldsymbol{q}_{\omega,k} \leftarrow \frac{\boldsymbol{q}_{\omega,k}}{\|\boldsymbol{q}_{\omega,k}\|}$
        \STATE $\|\boldsymbol{\omega}_k\|^2 \leftarrow \omega_{x,k}^2+\omega_{y,k}^2+\omega_{z,k}^2$
        \IF{$t_k \leq 3 \text{ seconds}$}
            \STATE $w_{a,k}=0.5$
        \ELSE
            \STATE $w_{a,k}=\alpha_ae^{-\beta_a\|\boldsymbol{\omega}_k\|^2}$
        \ENDIF
        \STATE $s_{a,k}=\boldsymbol{q}_{\omega,k}\cdot\boldsymbol{q}_{a,k}$
        \STATE $\Lambda_{a,k}=\{1+\sqrt{1-4w_{a,k}(1-w_{a,k})(1-s_{a,k}^2)}\}/{2}$
        \STATE $\boldsymbol{q}_{a\omega,k}\leftarrow\boldsymbol{q}_{\omega,k}+\{{w_{a,k}s_{a,k}/(\Lambda_{a,k}-w_{a,k}})\}\boldsymbol{q}_{a,k}$
        \STATE $\boldsymbol{q}_{a\omega,k}\leftarrow\frac{\boldsymbol{q}_{a\omega,k}}{\|\boldsymbol{q}_{a\omega,k}\|}$
        \STATE Calculate $\boldsymbol{q}_{gm,k}$ using (\ref{eq:calculation_of_qgm}).
        \IF{$t_k \leq 3 \text{ seconds}$}
            \STATE $w_{m,k}=0.5$
        \ELSE
            \STATE $e_{m,k}=\frac{m_k^{abs}-m_s}{m_s}$
            \STATE 
    $w_{m,k}=\alpha_m\exp\!\left(-\beta_a\|\boldsymbol{\omega}_k\|^2-\beta_me_{m,k}^2\right)$
        \ENDIF
        \STATE $s_{m,k}=\boldsymbol{q}_{a\omega,k}\cdot\boldsymbol{q}_{gm,k}$
        \STATE $\Lambda_{m,k}=\{1+\sqrt{1-4w_{m,k}(1-w_{m,k})(1-s_{m,k}^2)}\}/{2}$
        \STATE $\hat{\boldsymbol{q}}_k=\boldsymbol{q}_{a\omega,k}+\{w_{m,k}s_{m,k}/(\Lambda_{m,k}-w_{m,k})\}\boldsymbol{q}_{gm,k}$
        \STATE $\hat{\boldsymbol{q}}_k\leftarrow\frac{\hat{\boldsymbol{q}}_k}{\|\hat{\boldsymbol{q}}_k\|}$
    \ENDFOR
    \RETURN $\hat{\boldsymbol{q}}_k$
	\end{algorithmic}
\end{algorithm}
\end{figure}

\subsection{Pedestrian Dead Reckoning}
The walking trajectory is estimated by double integration of the acceleration in the global coordinate system. However, velocity and trajectory estimation errors accumulate due to accelerometer bias and measurement noise. To reduce these errors, the Zero Velocity Update (ZVU) algorithm is applied. The ZVU algorithm corrects the estimated velocity so that the velocity becomes zero during the midstance phase under the assumption that the foot velocity is zero during the midstance phase \cite{a21,a22}. The midstance phase is detected using a threshold-based algorithm based on the magnitude of the angular velocity \cite{a20}. If $\|\boldsymbol{\omega}_k\|\leq\theta_{ms}$, the midstance phase is detected, where $\theta_{ms}$ denotes the threshold for midstance phase detection \cite{a20}. According to \cite{a21,a22}, the bias error used in the ZVU algorithm is estimated by
\begin{align}
    \begin{bmatrix}
        0\\
        \boldsymbol{\epsilon}
    \end{bmatrix}
    =\frac{1}{T_s}\int_{t_s}^{t_e}\biggl(\hat{\boldsymbol{q}}_k\otimes
    \begin{bmatrix}
        0\\
        \boldsymbol{a}_k
    \end{bmatrix}
    \otimes \hat{\boldsymbol{q}}_k^*-
    \begin{bmatrix}
        \boldsymbol{0}_{3\times1}\\
        g
    \end{bmatrix}
    \biggr)
    dt
\end{align}
where $T_s=t_e-t_s$, $\boldsymbol{0}_{3\times1}$, $\hat{\boldsymbol{q}}_k^*$, and $g$ denote the duration from the beginning of toe-off to the end of heel strike, the zero vector of size $3\times1$, the conjugate of the quaternion $\hat{\boldsymbol{q}}_k$, and gravitational acceleration, respectively.$t_s$ and $t_e$ denote the time of the beginning of toe-off, and the time of the end of heel strike, respectively. When the foot is not in the midstance phase, the velocity $\boldsymbol{v}_k$ is estimated by
\begin{align}
    \begin{bmatrix}
        0\\
        \hat{\boldsymbol{v}}_k
    \end{bmatrix}
    =\int_{t_s}^{t_e} \biggl(\hat{\boldsymbol{q}}_k\otimes
    \begin{bmatrix}
        0\\
        \boldsymbol{a}_k
    \end{bmatrix}
    \otimes \hat{\boldsymbol{q}}_k^*-
    \begin{bmatrix}
        \boldsymbol{0}_{3\times1}\\
        g
    \end{bmatrix}-
    \begin{bmatrix}
        0\\
        \boldsymbol{\epsilon}
    \end{bmatrix}
    \biggr)
    dt
\end{align}
Finally, the walking trajectory $\boldsymbol{p}_k$ is estimated as
\begin{align}
    \hat{\boldsymbol{p}}_k=\int_{t_0}^{t_k}\hat{\boldsymbol{v}}_k dt
\end{align}

\section{Experiments}
\subsection{Experimental Setup}
\begin{figure}[htbp]
    \centering
    \includegraphics[width=90mm]{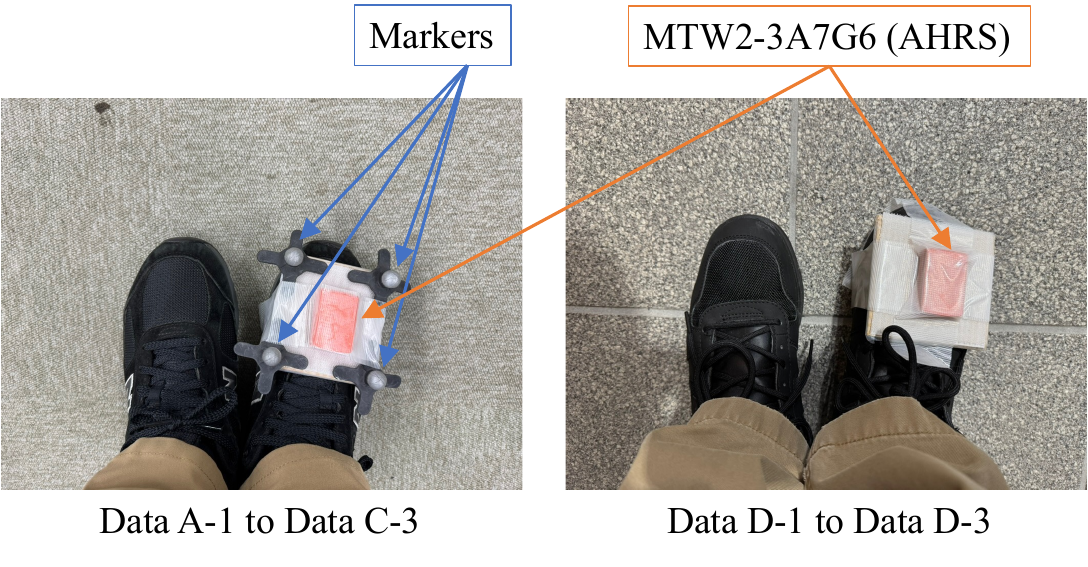}
    \caption{Mounted AHRS and markers on the foot}
    \label{fig:mounting_devices}
\end{figure}

The three experiments were conducted to verify the accuracy of the proposed QAACF and compare its computational cost with those of existing filters. Table \ref{table:Experimental_datasets} lists the walking datasets collected in the indoor experiments. MTW2-3A7G6, manufactured by Movella Inc. \cite{b20}, and OptiTrack, manufactured by NaturalPoint Inc. \cite{a23}, were used to collect Data A-1 to Data C-3, whereas only MTW2-3A7G6 was used to collect Data D-1 to Data D-3. The noise densities of the MTW2-3A7G6 for angular velocity, acceleration, and the magnetic field are 0.01 $deg/s/\sqrt{Hz}$, 200 $\mu g/\sqrt{Hz}$, and 0.2 $mGauss/\sqrt{Hz}$, respectively \cite{b20}. The sampling rate was set to 100 Hz. Fig. \ref{fig:mounting_devices} shows the mounting configuration of the AHRS sensor (MTW2-3A7G6) and the markers for the optical motion capture system (OptiTrack).\\
The proposed QAACF was compared with the existing KFs (EKF, Guo's FKF \cite{b1}, Wu's RMr-GDALKF \cite{b13}, and Yamagishi's KCKF \cite{a13}) and CFs (Wu's FCF \cite{b6}, Madgwick's algorithm \cite{b4}, and Mahony's algorithm \cite{b5}). For the KFs, the parameters were set to $\sigma_{\omega}^2=1.00\times10^{-3}$, $\sigma_{acc}^2=1.00\times10^{-2}$, $\sigma_{mag}^2=1.00\times10^{-2}$, and $\boldsymbol{P}_0=\boldsymbol{I}_{4\times4}$, where $\sigma_{\omega}^2$, $\sigma_{acc}^2$, and $\sigma_{mag}^2$ denote the noise variances of the angular velocity, acceleration, and magnetic field, respectively, and $\boldsymbol{P}_0$ denotes the initial error covariance matrix. For the proposed QAACF, the parameters were set to $\mu=0.15$, $\alpha_a=0.50$, $\beta_a=200.0$, $\alpha_m=0.05$, $\beta_m=200.0$, and $m_s=1.0$. For Wu's FCF, the parameters were set to $\gamma_a=\gamma_m=0.01$. In addition, the parameter of Madgwick's algorithm was set to $\beta=0.15$, whereas the parameters of Mahony's algorithm were set to $k_I=0.01$ and $k_P=2.0$.\\
Before attitude estimation, the gyroscope data were calibrated using static data collected for about ten minutes, while the magnetometer data were calibrated using ellipsoid calibration data.\\
All the filters were implemented using NumPy arrays in Python 3. However, only the calculation of $\boldsymbol{q}_{gm,k}$ in the proposed QAACF and Wu's FCF was implemented component-wise using scalar operations instead of NumPy array-based computations to reduce the computational cost. In addition, papers \cite{b5} and \cite{a24} were used as references for implementing Mahony's algorithm.\\

\begin{table*}[t]
\centering
\caption{Experimental datasets}
\label{table:Experimental_datasets}
\begin{tabular}{cccccc}
\hline
Data Name & Place & Walking Path & Devices Used & Walked Distance [m] & Measurement Time [s] \\
\hline
Data A-1 & Indoor & Straight & MTW2-3A7G6, OptiTrack & 2.94 & 24.60 \\
Data A-2 & Indoor & Straight & MTW2-3A7G6, OptiTrack & 2.79 & 21.15 \\
Data A-3 & Indoor & Straight & MTW2-3A7G6, OptiTrack & 2.91 & 20.46 \\
Data B-1 & Indoor & Rectangular path & MTW2-3A7G6, OptiTrack & 10.30 & 29.13 \\
Data B-2 & Indoor & Rectangular path & MTW2-3A7G6, OptiTrack & 9.77 & 27.89 \\
Data B-3 & Indoor & Rectangular path & MTW2-3A7G6, OptiTrack & 9.67 & 28.08 \\
Data C-1 & Indoor & Rectangular path $\times$ 10 & MTW2-3A7G6, OptiTrack & 93.48 & 100.73 \\
Data C-2 & Indoor & Rectangular path $\times$ 10 & MTW2-3A7G6, OptiTrack & 93.07 & 91.99 \\
Data C-3 & Indoor & Rectangular path $\times$ 10 & MTW2-3A7G6, OptiTrack & 93.04 & 92.23 \\
Data D-1 & Indoor & Walking long distance in a rectangular path & MTW2-3A7G6 & 354.45 & 333.76 \\
Data D-2 & Indoor & Walking long distance in a rectangular path & MTW2-3A7G6 & 354.066 & 335.75 \\
Data D-3 & Indoor & Walking long distance in a rectangular path & MTW2-3A7G6 & 354.802 & 342.95 \\
\hline
\end{tabular}
\end{table*}

\subsection{Experiment 1: Comparison of Attitude Estimation Accuracy}
\begin{figure}[!t]
    \centering
    \includegraphics[width=90mm]{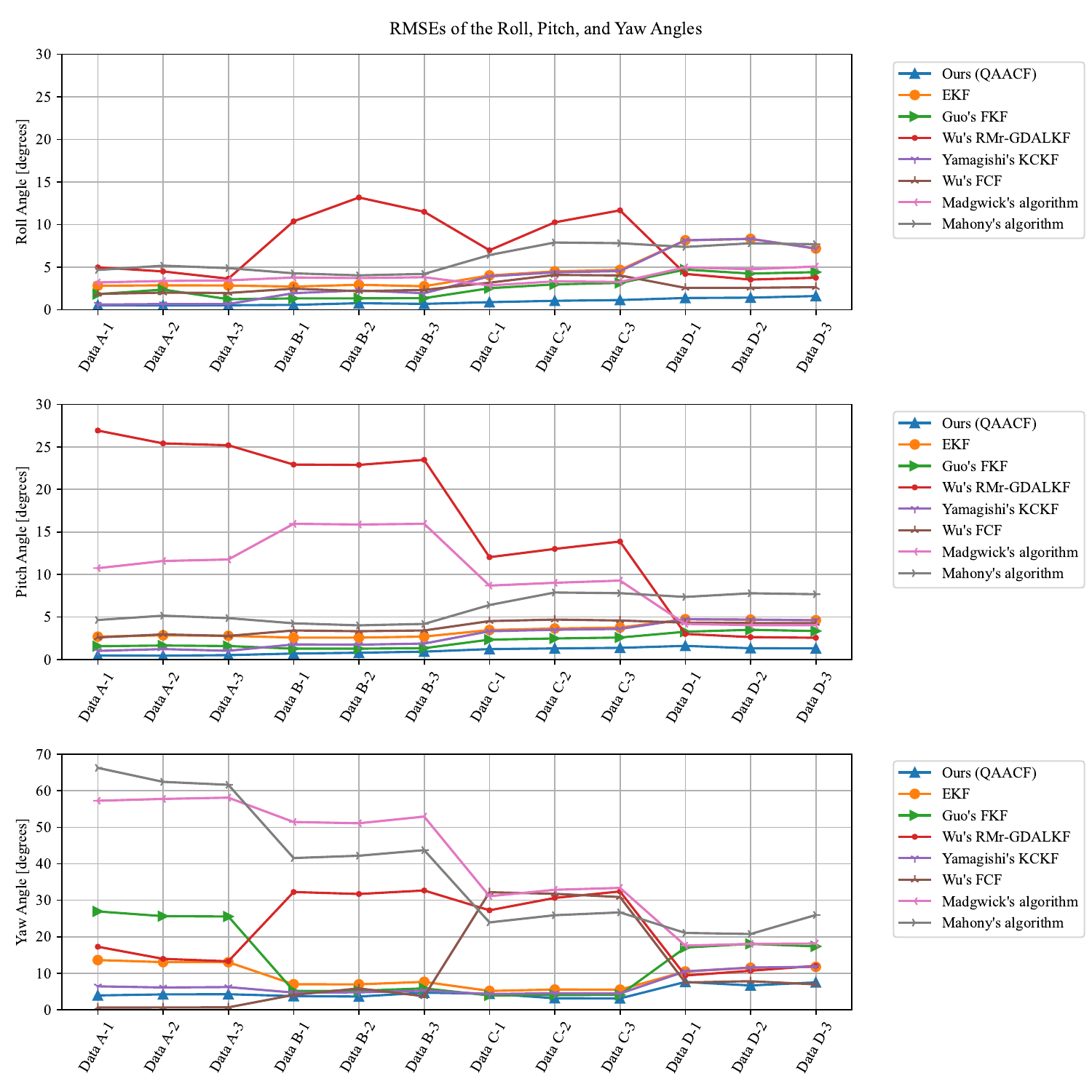}
    \caption{Line graph of the attitude RMSEs shown in Table \ref{table:RMSE_attitude}}
    \label{fig:oresen_RMSE_Eulers}
\end{figure}

\begin{figure}[!t]
    \centering
    \includegraphics[width=90mm]{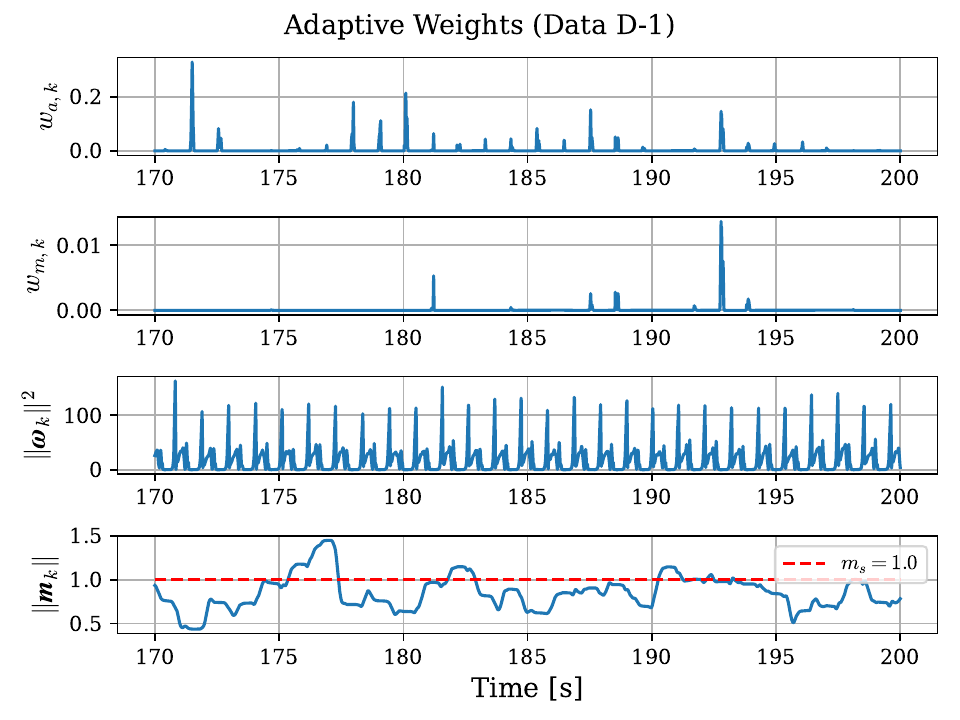}
    \caption{Time histories of the adaptive weights}
    \label{fig:adaptive_weights}
\end{figure}

\begin{figure*}[!t]
    \centering
    \includegraphics[width=180mm]{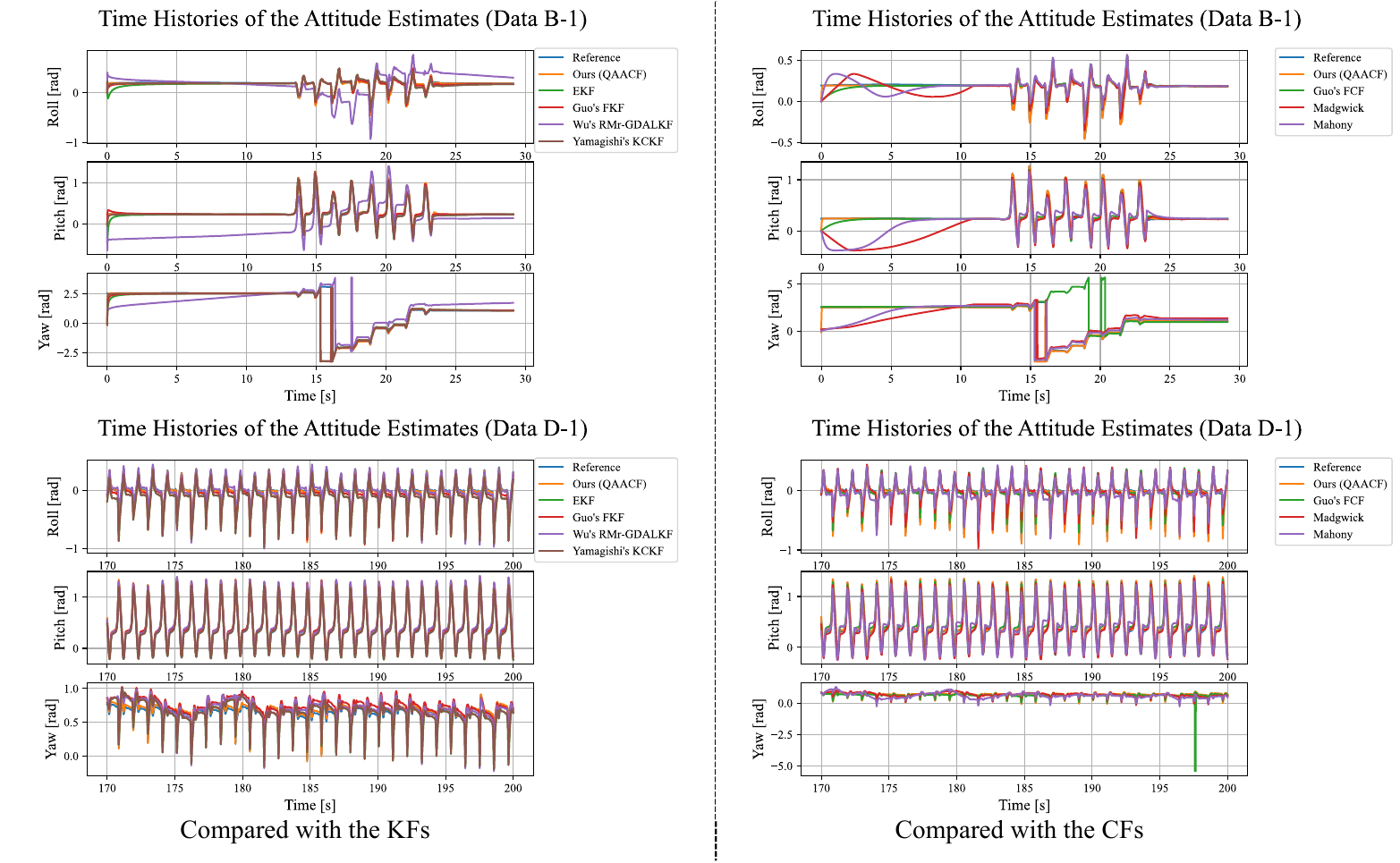}
    \caption{Time histories of the attitudes}
    \label{fig:attitude_waveforms}
\end{figure*}

This experiment was conducted to evaluate the attitude estimation accuracy of the proposed QAACF. The attitude estimates obtained from the Xsens Kalman Filter (XKF3hm) \cite{b20} were used as the reference because they provide reliable attitude estimates. The root-mean-square error (RMSE), defined by (\ref{eq:RMSE_Euler}), was used to evaluate the accuracy of each attitude estimation algorithm.
\begin{align}
    \label{eq:RMSE_Euler}
    \text{RMSE}&=\sqrt{\frac{1}{n}\sum_{k=0}^{n-1}[\arg\{e^{(\hat{\theta}_k-\theta_k)j}\}]^2}\\
    &=\sqrt{
\frac{1}{n}\sum_{k=0}^{n-1}\left[\operatorname{atan2}\left(\sin(\hat{\theta}_k-\theta_k), \cos(\hat{\theta}_k-\theta_k)\right)\right]^2}
\end{align}
Here, $j$, $\hat{\theta}$, and $\theta$ denote the imaginary unit, the Euler angle (roll, pitch, or yaw) estimated by each attitude estimation algorithm, and the corresponding reference Euler angle estimated by XKF3hm, respectively. The offset between the yaw angle estimated by each algorithm and that estimated by XKF3hm was removed before calculating the yaw RMSE. Table \ref{table:RMSE_attitude} summarizes the average and standard deviation (SD) of the RMSEs of the Euler angles, and Fig. \ref{fig:oresen_RMSE_Eulers} presents a line graph summarizing the results in Table \ref{table:RMSE_attitude}.\\
According to the experimental results, the proposed QAACF achieved the lowest attitude RMSEs (roll, pitch, and yaw) among the compared KFs and CFs. Two factors may account for this result. First, the adaptive weights were set to 0.5 to accelerate convergence, and the attitude estimates converged rapidly. Second, the adaptive weighting scheme effectively improved the attitude estimation accuracy.\\
In addition, Fig. \ref{fig:adaptive_weights} and Fig. \ref{fig:attitude_waveforms} show the time histories of the adaptive weights and the estimated attitudes, respectively. The time intervals during which the acceleration measurements are reliable are very short because the midstance phase has a very short duration. Furthermore, $w_{m,k}$ remains very low and is almost zero for most of the time because the magnetic field is unstable and subject to magnetic disturbances.

\begin{table*}[!t]
    \centering
    \caption{Average and standard deviation of the RMSE of Euler angles (Values in parentheses denote SD; Unit: degrees)}
    \label{table:RMSE_attitude}
    \renewcommand{\arraystretch}{1.2}
    \begin{tabular}{lccc}
    \toprule
    Method & Roll (degrees) & Pitch (degrees) & Yaw (degrees)\\
    \midrule
    Ours (QAACF) & 0.914 (0.390) & 1.016 (0.404) & 4.711 (1.603) \\
    EKF & 4.478 (2.183) & 3.437 (0.861) & 9.248 (3.280) \\
    Guo's FKF & 2.617 (1.287) & 2.197 (0.848) & 13.214 (9.483) \\
    Wu's RMr-GDALKF & 7.379 (3.726) & 16.164 (9.531) & 21.947 (9.889) \\
    Yamagishi's KCKF & 3.710 (2.874) & 2.774 (1.471) & 6.688 (2.860) \\
    Wu's FCF & 2.651 (0.746) & 3.780 (0.763) & 11.028 (12.677) \\
    Madgwick's Algorithm & 3.794 (0.739) & 10.103 (4.465) & 39.975 (16.52) \\
    Mahony's Algorithm & 6.016 (1.619) & 12.618 (3.233) & 38.504 (17.112) \\
    \bottomrule
    \end{tabular}
\end{table*}

\subsection{Experiment 2: Comparison of PDR Estimation Accuracy}
This experiment was conducted to evaluate the PDR estimation accuracy of the proposed QAACF. The walking trajectories obtained from the OptiTrack \cite{a23} were used as the reference trajectories for Data A-1 to Data C-3, whereas the walking trajectories estimated by the PDR with the XKF3hm \cite{b20} were used as the reference trajectories for Data D-1 to Data D-3.

\begin{figure}[!t]
    \centering
    \includegraphics[width=100mm]{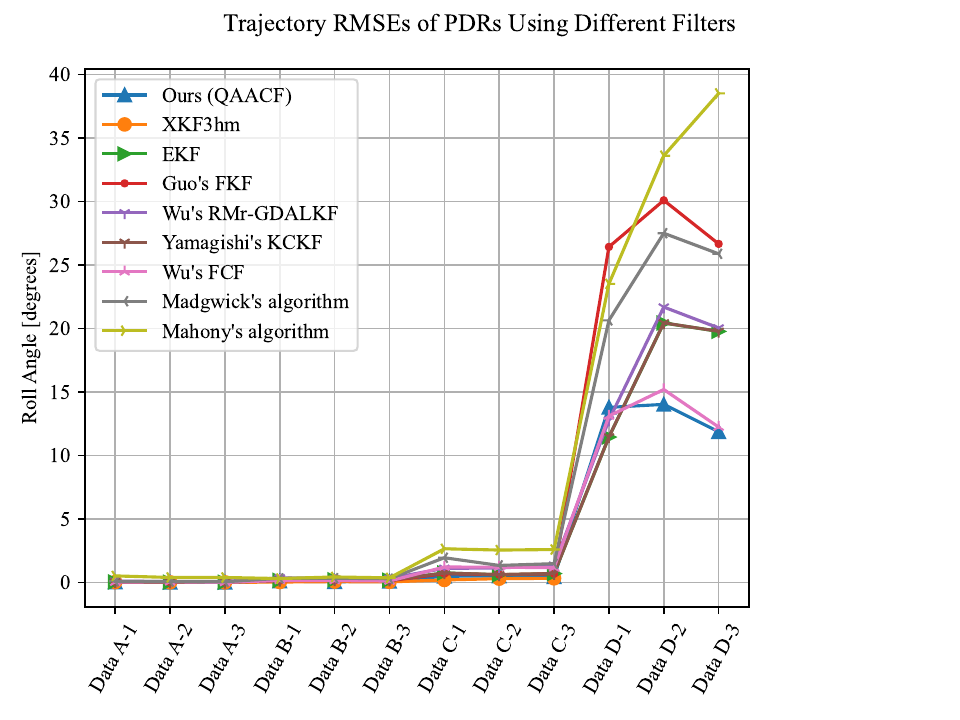}
    \caption{Line graph of the trajectory RMSEs}
    \label{fig:oresen_RMSE_trajectory}
\end{figure}

\begin{figure}[!t]
    \centering
    \includegraphics[width=100mm]{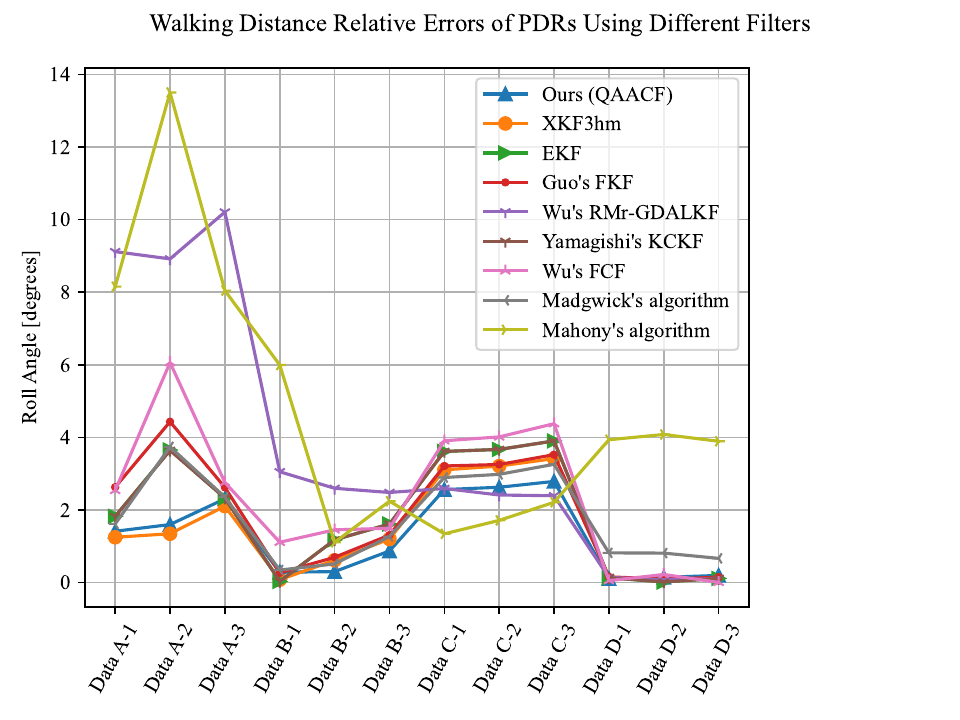}
    \caption{Line graph of the relative errors of the estimated walking distances}
    \label{fig:oresen_RE_distance}
\end{figure}

\begin{figure*}[!t]
    \centering
    \includegraphics[width=163mm]{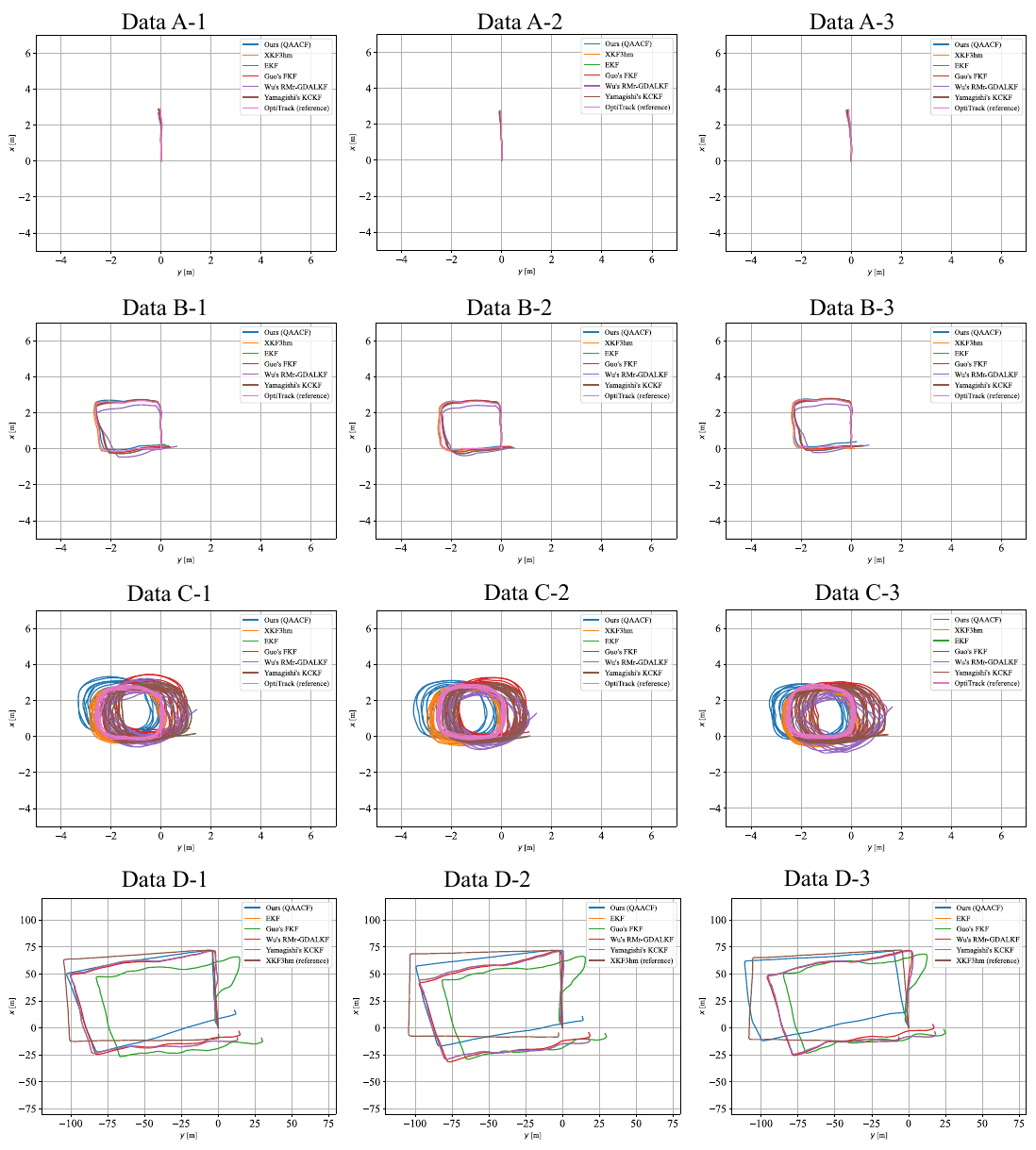}
    \caption{Walking trajectories estimated by PDR using QAACF and the Kalman filters}
    \label{fig:trajectory_KFs}
\end{figure*}

\begin{figure*}[!t]
    \centering
    \includegraphics[width=163mm]{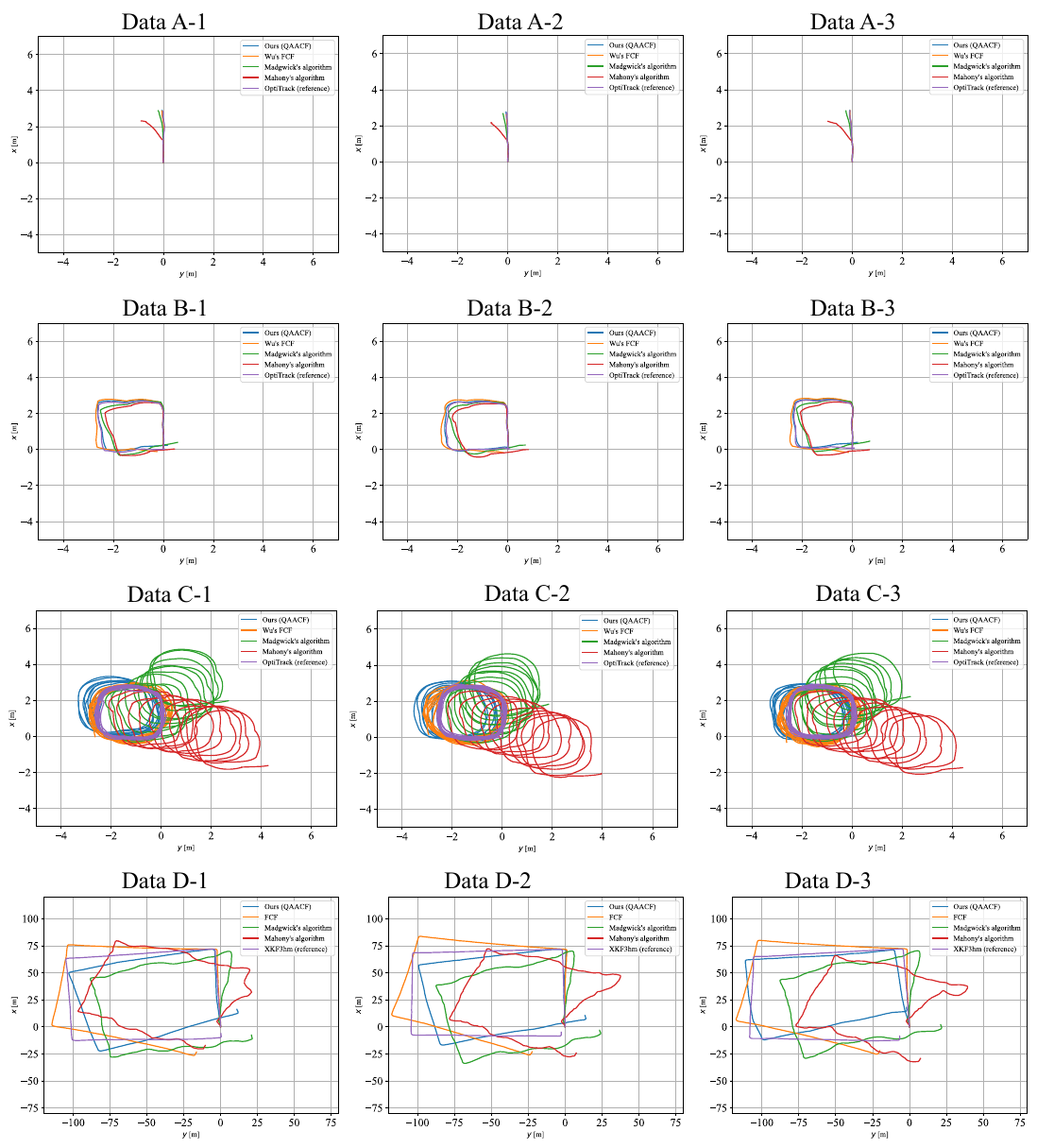}
    \caption{Walking trajectories estimated by PDR using QAACF and the Complementary filters}
    \label{fig:trajectory_CFs}
\end{figure*}

The root-mean-square error (RMSE), defined in (\ref{eq:RMSE_trajectory}), was used to evaluate the trajectory estimation accuracy of the PDR with each attitude estimation algorithm.
\begin{align}
    \label{eq:RMSE_trajectory}
    \mathrm{RMSE}
    =\sqrt{\frac{1}{n}\sum_{k=0}^{n-1}|\hat{\boldsymbol{p}}_k-\boldsymbol{p}_k|^2}
\end{align}
Here, $\hat{\boldsymbol{p}}_k$ and $\boldsymbol{p}_k$ denote the walking trajectory estimated by the PDR with each attitude estimation algorithm and the corresponding reference trajectory obtained from the OptiTrack or the PDR with the XKF3hm, respectively. In addition, the relative error (RE), defined in (\ref{eq:relative_error}), was used to evaluate the accuracy of walking distance estimation by the PDR with each attitude estimation algorithm.
\begin{align}
    \label{eq:relative_error}
    \mathrm{RE}
    =\frac{|\hat{d}-d|}{d}\times100
\end{align}
Here, $\hat{d}$ and $d$ denote the walking distance estimated by the PDR with each attitude estimation algorithm and the corresponding reference walking distance obtained from the OptiTrack or the PDR with the XKF3hm, respectively.\\
Table \ref{table:trajectory_rmse} presents the trajectory RMSEs, and Fig. \ref{fig:oresen_RMSE_trajectory} presents the corresponding line graph. Table \ref{table:relative_error_distance} presents the relative errors of the estimated walking distances, and Fig. \ref{fig:oresen_RE_distance} presents the corresponding line graph. Additionally, Figs. \ref{fig:trajectory_KFs} and \ref{fig:trajectory_CFs} show the walking trajectories estimated by the PDR using QAACF and KFs and those estimated by the PDR using QAACF and CFs, respectively.
According to the experimental results, the PDR using QAACF generally achieves low trajectory RMSEs and relative errors of the estimated walking distance compared to the PDRs using the other attitude estimation filters. This is considered the result of the higher accuracy of the QAACF's attitude estimation.

\begin{table*}[!t]
\centering
\caption{Trajectory RMSEs (Unit: m)}
\label{table:trajectory_rmse}
\begin{tabular}{lccccccccc}
\toprule
 & Ours (QAACF) & XKF3hm & EKF & Guo's FKF & Wu's RMr-GDALKF & Yamagishi's KCKF & Wu's FCF & Madgwick & Mahony \\
\midrule
Data A-1 & 0.039 & 0.039 & 0.053 & 0.050 & 0.107 & 0.049 & 0.036 & 0.102 & 0.525 \\
Data A-2 & 0.020 & 0.020 & 0.036 & 0.034 & 0.094 & 0.034 & 0.065 & 0.078 & 0.404 \\
Data A-3 & 0.014 & 0.015 & 0.028 & 0.042 & 0.093 & 0.024 & 0.012 & 0.063 & 0.402 \\
Data B-1 & 0.110 & 0.043 & 0.178 & 0.146 & 0.372 & 0.176 & 0.169 & 0.335 & 0.317 \\
Data B-2 & 0.065 & 0.076 & 0.215 & 0.168 & 0.299 & 0.213 & 0.161 & 0.344 & 0.443 \\
Data B-3 & 0.093 & 0.055 & 0.188 & 0.148 & 0.327 & 0.186 & 0.122 & 0.303 & 0.372 \\
Data C-1 & 0.499 & 0.207 & 0.759 & 0.700 & 1.124 & 0.756 & 1.225 & 1.956 & 2.661 \\
Data C-2 & 0.516 & 0.310 & 0.618 & 0.627 & 1.145 & 0.615 & 1.215 & 1.350 & 2.560 \\
Data C-3 & 0.447 & 0.336 & 0.710 & 0.691 & 1.373 & 0.706 & 1.167 & 1.490 & 2.602 \\
Data D-1 & 13.792 & -- & 11.459 & 26.424 & 12.791 & 11.467 & 13.110 & 20.643 & 23.515 \\
Data D-2 & 14.028 & -- & 20.433 & 30.092 & 21.689 & 20.442 & 15.199 & 27.502 & 33.603 \\
Data D-3 & 11.875 & -- & 19.776 & 26.663 & 20.042 & 19.779 & 12.245 & 25.881 & 38.510 \\
\midrule
\begin{tabular}[c]{@{}l@{}}Average\\(SD)\end{tabular}
& \begin{tabular}[c]{@{}c@{}}3.458\\(5.918)\end{tabular}
& \begin{tabular}[c]{@{}c@{}}0.122\\(0.127)\end{tabular}
& \begin{tabular}[c]{@{}c@{}}4.538\\(7.945)\end{tabular}
& \begin{tabular}[c]{@{}c@{}}7.149\\(12.442)\end{tabular}
& \begin{tabular}[c]{@{}c@{}}4.955\\(8.235)\end{tabular}
& \begin{tabular}[c]{@{}c@{}}4.537\\(7.949)\end{tabular}
& \begin{tabular}[c]{@{}c@{}}3.727\\(5.958)\end{tabular}
& \begin{tabular}[c]{@{}c@{}}6.671\\(10.982)\end{tabular}
& \begin{tabular}[c]{@{}c@{}}8.826\\(14.308)\end{tabular} \\
\bottomrule
\end{tabular}
\end{table*}

\begin{table*}[!t]
\centering
\caption{Relative errors of the estimated walking distance (Unit: \%)}
\label{table:relative_error_distance}
\begin{tabular}{lccccccccc}
\toprule
 & Ours (QAACF) & XKF3hm & EKF & Guo's FKF & Wu's RMr-GDALKF & Yamagishi's KCKF & Wu's FCF & Madgwick & Mahony\\
\midrule
Data A-1 & 1.415 & 1.249 & 1.827 & 2.629 & 9.118 & 1.824 & 2.540 & 1.616 & 8.153 \\
Data A-2 & 1.599 & 1.346 & 3.653 & 4.432 & 8.918 & 3.641 & 6.065 & 3.740 & 13.503 \\
Data A-3 & 2.308 & 2.106 & 2.328 & 2.623 & 10.209 & 2.330 & 2.759 & 2.355 & 8.046 \\
Data B-1 & 0.285 & 0.080 & 0.035 & 0.269 & 3.054 & 0.034 & 1.112 & 0.346 & 5.994 \\
Data B-2 & 0.303 & 0.615 & 1.179 & 0.697 & 2.598 & 1.179 & 1.454 & 0.515 & 1.110 \\
Data B-3 & 0.871 & 1.201 & 1.614 & 1.310 & 2.483 & 1.614 & 1.498 & 1.269 & 2.242 \\
Data C-1 & 2.563 & 3.107 & 3.614 & 3.213 & 2.593 & 3.614 & 3.908 & 2.895 & 1.344 \\
Data C-2 & 2.627 & 3.210 & 3.669 & 3.254 & 2.413 & 3.669 & 4.014 & 2.983 & 1.714 \\
Data C-3 & 2.788 & 3.417 & 3.903 & 3.525 & 2.393 & 3.904 & 4.378 & 3.259 & 2.214 \\
Data D-1 & 0.112 & -- & 0.133 & 0.159 & 0.145 & 0.133 & 0.054 & 0.819 & 3.936 \\
Data D-2 & 0.145 & -- & 0.018 & 0.097 & 0.074 & 0.018 & 0.224 & 0.813 & 4.081 \\
Data D-3 & 0.198 & -- & 0.114 & 0.147 & 0.110 & 0.114 & 0.006 & 0.668 & 3.894 \\
\midrule
\begin{tabular}[c]{@{}l@{}}Average\\(SD)\end{tabular}
& \begin{tabular}[c]{@{}c@{}}1.268\\(1.081)\end{tabular}
& \begin{tabular}[c]{@{}c@{}}1.815\\(1.206)\end{tabular}
& \begin{tabular}[c]{@{}c@{}}1.841\\(1.573)\end{tabular}
& \begin{tabular}[c]{@{}c@{}}1.863\\(1.580)\end{tabular}
& \begin{tabular}[c]{@{}c@{}}3.676\\(3.635)\end{tabular}
& \begin{tabular}[c]{@{}c@{}}1.840\\(1.572)\end{tabular}
& \begin{tabular}[c]{@{}c@{}}2.334\\(1.946)\end{tabular}
& \begin{tabular}[c]{@{}c@{}}1.773\\(1.209)\end{tabular}
& \begin{tabular}[c]{@{}c@{}}4.686\\(3.683)\end{tabular} \\
\bottomrule
\end{tabular}
\end{table*}

\subsection{Experiment 3: Comparison of Computation Time Among Attitude Estimation Filters}
\begin{figure*}[t]
    \centering
    \includegraphics[width=160mm]{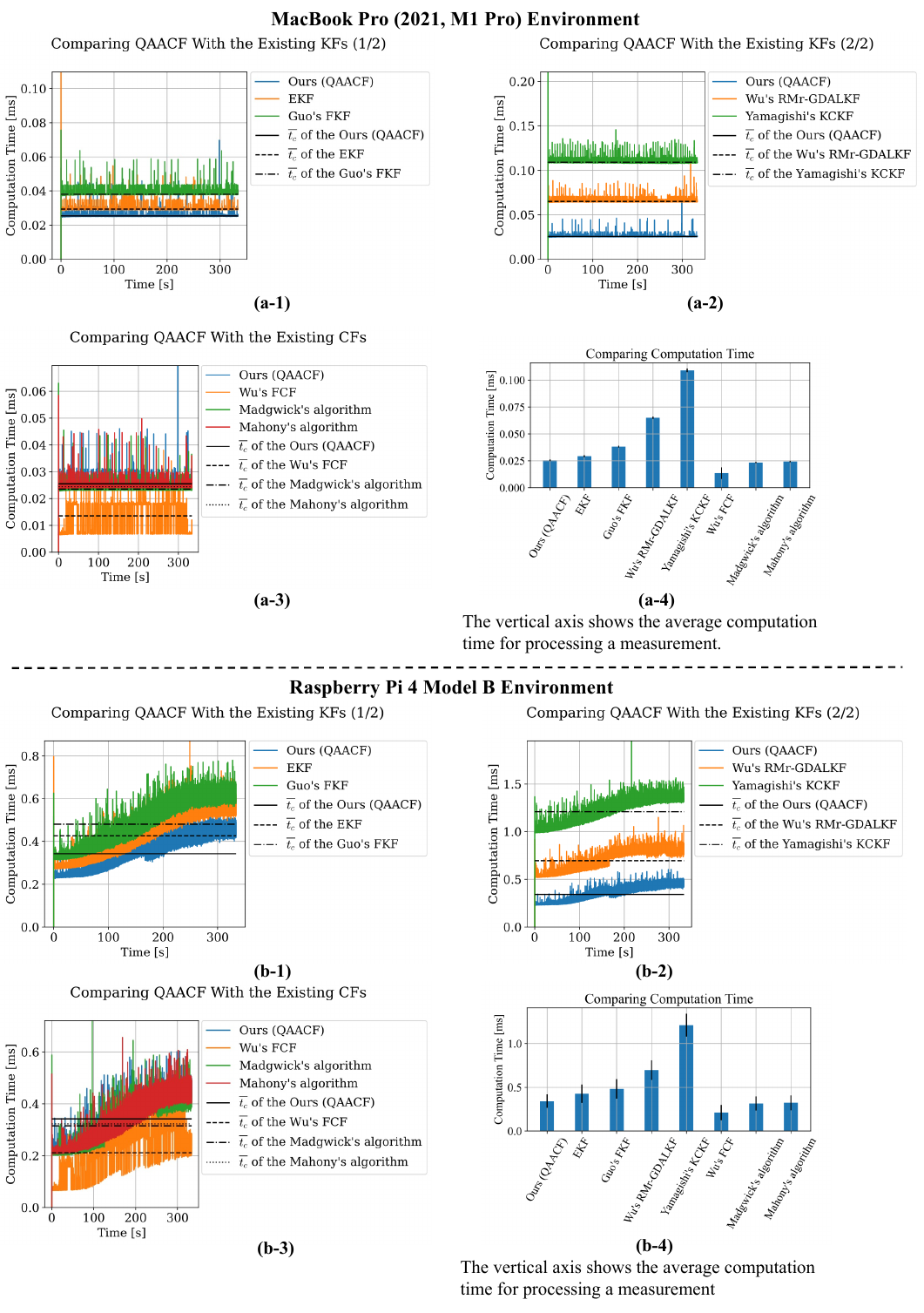}
    \caption{Measured computation times of each attitude estimation algorithm}
    \label{fig:computation_time}
\end{figure*}

This experiment was conducted to compare the computation time of the proposed QAACF with those of other attitude estimation filters. The computation time was measured in two different computing environments: a high-performance computing environment and a low-cost computing environment. A MacBook Pro (M1 Pro, 2021) \cite{b38} was used as the high-performance computing environment, whereas a Raspberry Pi 4 Model B \cite{b39} was used as the low-cost computing environment. Data D-1 was used to measure computation time. Fig. \ref{fig:computation_time} shows the measured computation times of each attitude estimation algorithm. In addition, Tables \ref{table:computation_time_mac} and \ref{table:computation_time_raspi} show the corresponding average computation times ($\overline{t_c}$) and standard deviations measured on a MacBook Pro (2021, M1 Pro) and a Raspberry Pi 4 Model B, respectively. $\overline{t_c}$ represents the average computation time for processing one set of acceleration, angular velocity, and magnetic field measurements. According to the experimental results, the computation time of the proposed QAACF is lower than those of the KFs. Therefore, QAACF is suitable for low-cost computing environments.

\begin{table}[!t]
    \centering
    \caption{Average computation time and standard deviation on a MacBook Pro (2021, M1 Pro)}
    \label{table:computation_time_mac}
    \begin{tabular}{lcc}
    \toprule
    Algorithm & $\overline{t}_c$ [ms] & SD [ms] \\
    \midrule
    Ours (QAACF)          & $2.55\times10^{-2}$ & $7.19\times10^{-4}$ \\
    EKF                   & $2.93\times10^{-2}$ & $8.07\times10^{-4}$ \\
    Guo's FKF             & $3.81\times10^{-2}$ & $8.34\times10^{-4}$ \\
    Wu's RMr GDALKF       & $6.50\times10^{-2}$ & $1.09\times10^{-3}$ \\
    Yamagishi's KCKF      & $1.09\times10^{-1}$ & $1.63\times10^{-3}$ \\
    Wu's FCF              & $1.35\times10^{-2}$ & $5.39\times10^{-3}$ \\
    Madgwick's algorithm  & $2.35\times10^{-2}$ & $6.04\times10^{-4}$ \\
    Mahony's algorithm    & $2.44\times10^{-2}$ & $6.29\times10^{-4}$ \\
    \bottomrule
    \end{tabular}
\end{table}

\begin{table}[!t]
    \centering
    \caption{Average computation time and standard deviation on a Raspberry Pi 4 Model B}
    \label{table:computation_time_raspi}
    \begin{tabular}{lcc}
    \toprule
    Algorithm & $\overline{t}_c$ [ms] & SD [ms] \\
    \midrule
    Ours (QAACF)          & $3.42\times10^{-1}$ & $7.73\times10^{-2}$ \\
    EKF                   & $4.26\times10^{-1}$ & $1.05\times10^{-1}$ \\
    Guo's FKF             & $4.81\times10^{-1}$ & $1.11\times10^{-1}$ \\
    Wu's RMr GDALKF       & $6.95\times10^{-1}$ & $1.12\times10^{-1}$ \\
    Yamagishi's KCKF      & $1.21$              & $1.32\times10^{-1}$ \\
    Wu's FCF              & $2.11\times10^{-1}$ & $8.86\times10^{-2}$ \\
    Madgwick's algorithm  & $3.15\times10^{-1}$ & $8.10\times10^{-2}$ \\
    Mahony's algorithm    & $3.23\times10^{-1}$ & $8.57\times10^{-2}$ \\
    \bottomrule
    \end{tabular}
\end{table}

\clearpage
\clearpage
\section{Conclusion}

In this paper, a new complementary filter named QAACF was proposed. QAACF combines two quaternions more rigorously than conventional linear interpolation because it is derived from Markley’s quaternion averaging method. The proposed QAACF adaptively adjusts the weights to improve the estimation accuracy. According to the experimental results, the proposed QAACF achieves high estimation accuracy for walking in indoor environments while maintaining a low computational cost.\\
However, the proposed QAAFC does not necessarily achieve high accuracy in all cases and QAACF has some limitations. First, QAACF requires a high-accuracy gyroscope to achieve high estimation accuracy. In addition, accurate accelerometer and magnetometer measurements are also required to achieve high estimation accuracy. Second, QAACF requires gyroscope calibration using static data to achieve high estimation accuracy. In addition, the magnetometer should be calibrated using elliptical calibration data. Third, it was not verified whether QAACF achieves high estimation accuracy under highly dynamic conditions, such as running. Fourth, the attitude estimation RMSEs were evaluated using the attitude estimates obtained by XKF3hm, but they are not the actual ground truth.\\
In future work, we need to verify the attitude estimation accuracy using more reliable reference values. Additionally, we need to verify the accuracy of QAACF under highly dynamic conditions, such as running. In addition, we need to compare the accuracy of QAACF with those of other attitude estimation filters in outdoor environments because the magnetic field is generally more stable outdoors than indoors.

\bibliography{refs}
\bibliographystyle{IEEEtran}

\begin{IEEEbiography}[{\includegraphics[width=1in,height=1.25in,clip,keepaspectratio]{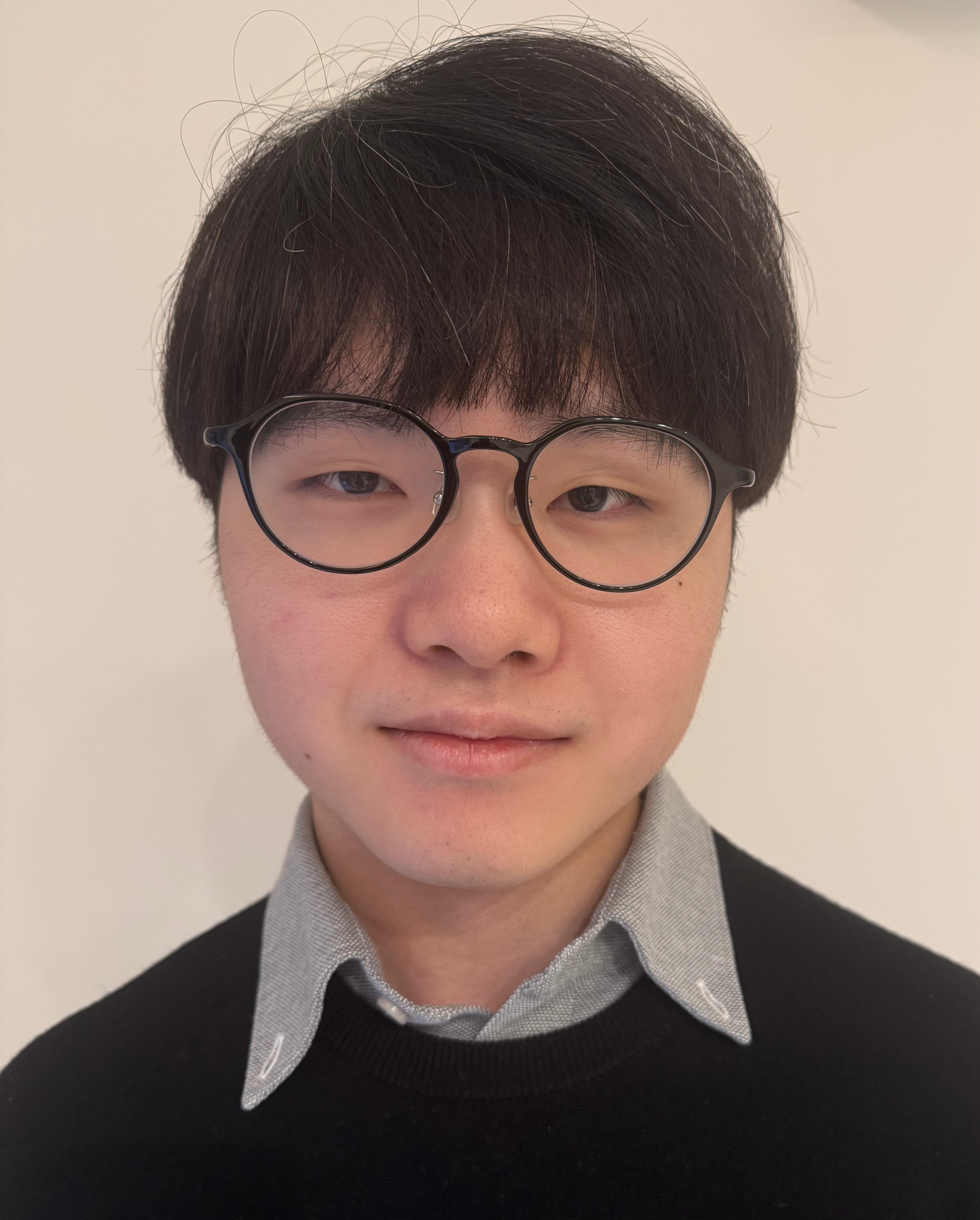}}]{Shunsei Yamagishi} received the B.S. degree and the M.S. degree in computer science and engineering from The University of  Aizu, Japan, in 2022 and 2024, respectively. He is currently working toward the Ph.D. degree in the Graduate School of Computer Science and Engineering, The University of Aizu, Japan. His research interests include algorithms for the attitude estimation for the Attitude and Heading Reference System, Pedestrian Dead Reckoning, signal processing, and sensor fusion methods.
\end{IEEEbiography}

\begin{IEEEbiography}[{\includegraphics[width=1in,height=1.25in,clip,keepaspectratio]{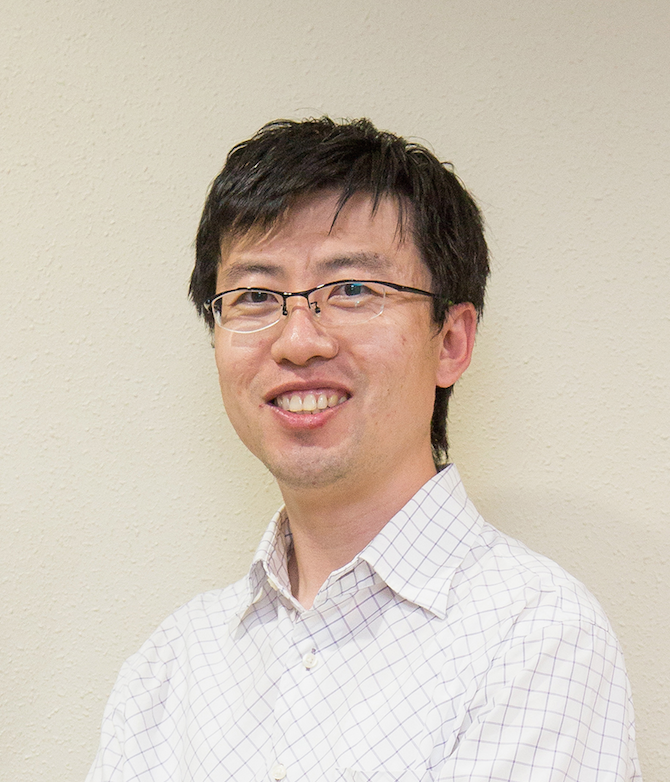}}]{Lei Jing} (M’12) received his Ph.D. degree in computer science and engineering from the University of Aizu, Japan, in 2008. He is currently a Senior Associate Professor at the School of Computer Science and Engineering, University of Aizu. His research interests include human position, posture, and motion tracking, soft circuit design, and the tactile internet. The applications of his work encompass human activity abnormality detection, sign language recognition, and human-robot interaction. He has published over 120 papers and holds six patents in related areas.
\end{IEEEbiography}

\vfill

\end{document}